%% file: main.tex
\PassOptionsToPackage{hyphens}{url}
\documentclass[10pt,twocolumn,letterpaper]{article}
\usepackage{cvpr}              
\usepackage{mdframed}
\usepackage{booktabs}
\usepackage{xspace}
\usepackage{amsmath}
\usepackage{multirow}
\usepackage[table]{xcolor}
\usepackage{array}
\usepackage{ragged2e}
\usepackage{adjustbox}
\usepackage{enumitem}
\usepackage{placeins}

\usepackage[accsupp]{axessibility} 

\newcommand{\token}[1]{\texttt{<#1>}}
\newcommand{\sot}{\token{startoftext}\xspace}
\newcommand{\eot}{\token{eot}\xspace}
\newcommand{\pad}{\token{pad}\xspace}

\newcommand{\bp}{\mathbf{v}^{\mathbf{pad}}}

\newcommand{\bt}{\mathbf{v}^{\mathbf{pr}}}
\newcommand{\bS}{\mathbf{v}^{\mathbf{sot}}}
\newcommand{\bE}{\mathbf{v}^{\mathbf{eot}}}
\newcommand{\bz}{\mathbf{0}}

\definecolor{cvprblue}{rgb}{0.21,0.49,0.74}
\usepackage[pagebackref,breaklinks,colorlinks,allcolors=cvprblue]{hyperref}

\def\paperID{} 
\def\confName{CVPR}
\def\confYear{2026}

\title{Memorization In Stable Diffusion Is Unexpectedly Driven by CLIP Embeddings}
\renewcommand{\thefootnote}{\fnsymbol{footnote}}
\author{
Bumjun Kim \qquad Albert No\footnotemark[2]\\
Department of Artificial Intelligence, Yonsei University\\
{\tt\small \{quasar529, albertno\}@yonsei.ac.kr}
}

\begin{document}
\maketitle
\footnotetext[2]{Corresponding author.}
\renewcommand{\thefootnote}{\arabic{footnote}}
\begin{abstract}
Understanding how textual embeddings contribute to memorization in text-to-image diffusion models is crucial for both interpretability and safety.
This paper investigates an unexpected behavior of CLIP embeddings in Stable Diffusion, revealing that the model disproportionately relies on specific embeddings.
We categorize input tokens as \sot, \token{prompt}, \texttt{<endoftext>} and \pad with corresponding embeddings $\bS, \bt, \bE, \bp$.
We discover that $\bt$ contribute minimally to generation in memorized cases.
In contrast, $\bp$ strongly affect memorization due to their structural duplication of $\bE$, the only embedding explicitly optimized during CLIP training.
This duplication unintentionally amplifies the influence of $\bE$, causing the model to over-rely on it, thereby driving memorization.
Based on these observations, we propose two simple yet effective inference-time mitigation strategies:
(1) Replacing the tokenizer’s default \pad from \eot to the \texttt{!} token before embedding, and masking the $\bE$; (2) Partial masking of $\bp$.
Both suppress memorization without degrading quality, and are readily deployable without prior detection.\footnotemark
\end{abstract}
\footnotetext{Code is available \href{https://github.com/quasar529/sd-clip-mem}{here}.}
\section{Introduction}
\label{intro}
Diffusion-based image generation models~\citep{ho2020denoising,songdenoising,song2021maximum,saharia2022photorealistic}  have seen rapid advancements, 
with Stable Diffusion~\citep{rombach2022high, podellsdxl} standing out for its efficiency, scalability, and widespread adoption.
However, alongside its success, Stable Diffusion exhibits a critical privacy risk known as memorization,
a phenomenon where the model reproduces images from its training set, 
raising concerns about privacy and copyright violations \citep{orrick2023andersen,jiang2023ai}.
Recent studies~\citep{somepalli2023diffusion, carlini2023extracting,webster2023reproducible} have demonstrated that Stable Diffusion can reconstruct training images,
revealing that large-scale diffusion models trained on extensive datasets inadvertently memorize and replicate private or copyrighted content.
This issue has drawn increasing scrutiny, leading to research efforts focused on identifying, detecting, 
and mitigating memorization~\citep{somepalli2023understanding, ren2024unveiling, daras2024ambient,darasconsistent, chen2024towards, hintersdorf2024finding,ross2024geometric, wen2024detecting, chenexploring,jiangimage}.

Memorization is often linked to dataset duplication~\citep{carlini2023extracting}, leading to efforts to filter redundant samples for mitigation.
Other studies focused on detecting trigger tokens~\citep{wen2024detecting} or analyzing cross-attention patterns~\citep{ren2024unveiling, chenexploring} to understand and mitigate memorization.
Although these approaches provide valuable insights into diffusion model behavior, 
they primarily focus on internal model dynamics. 
The role of the text embedding space, the interface through which prompts guide image generation, remains largely underexplored.
To address this overlooked aspect, we conduct a detailed investigation of the embedding space in Stable Diffusion.

\begin{table}[h]
\centering
\begin{tabular}{ll}
\toprule
\textbf{Token (Category: Sequence)} & \textbf{Embedding} \\
\midrule
\texttt{<startoftext>: <sot>} & $\bS$ \\
\texttt{<prompt>: <pr>$_1$,\dots,<pr>$_n$} & $\bt_1, …, \bt_n$ \\
\texttt{<endoftext>: <eot>} & $\bE$ \\
\texttt{<pad>: <pad>$_1$,\dots,<pad>$_d$} & $\bp_1, …, \bp_d$ \\
\bottomrule
\end{tabular}
\caption{
Text embedding vectors consist of four types.}
\label{tab:notation}
\vspace{-0.2em}
\end{table}

In this paper, we investigate how different components of textual embedding, particularly prompt and padding embeddings, contribute to memorization in text-to-image diffusion models.
To avoid ambiguity, we distinguish between tokens and their corresponding embedding vectors, as defined in Table~\ref{tab:notation}.
Given a fixed-length input of $L = n + d + 2$ tokens (typically 77), the text embedding is given by:
\[
\mathbf{Emb} = \big[\bS, \bt_1, \dots, \bt_n, \bE, \bp_1, \dots, \bp_d\big]
\]
where $\bS$ and $\bE$ denote the embedding of the \texttt{<sot>} and \eot, respectively; $\bt_i$ and $\bp_i$ denote the embeddings of the \texttt{<pr>$_i$} and \texttt{<pad>$_i$} .

\begin{figure*}[t]
    \centering
    \includegraphics[width=0.7\textwidth]{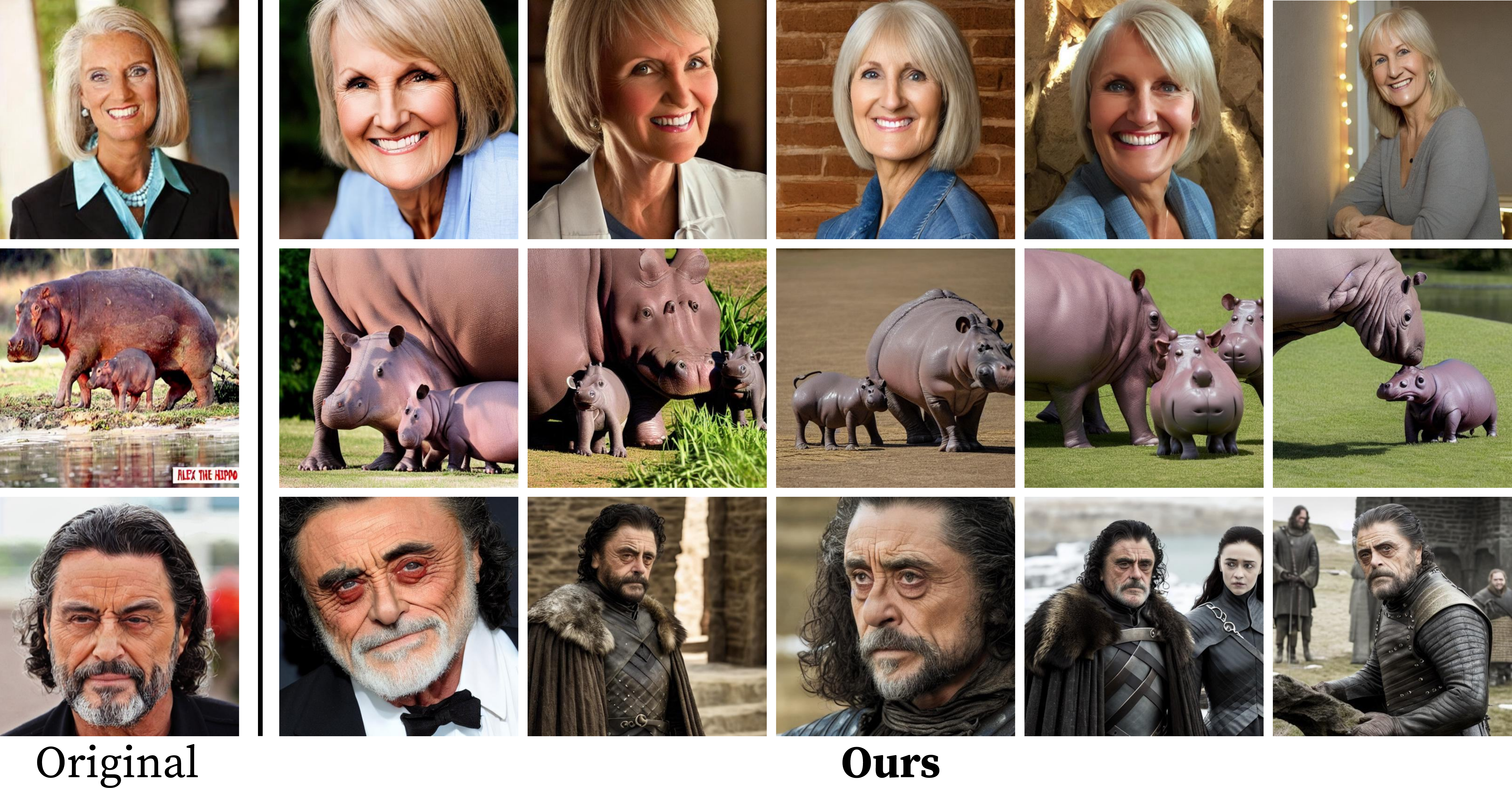}
\caption{
\textbf{Mitigating memorization via \pad replacement and $\bE$ masking.}  
We identify \pad embeddings ($\bp$) as a major contributor to memorization, as they are implicit duplications of the \texttt{<endoftext>} embedding ($\bE$). 
To mitigate this, we replace the tokenizer's default \pad (\eot) with the \texttt{!} token, and mask $\bE$.
This method effectively suppresses memorization while preserving image quality and prompt alignment. 
Each row shows results from the same prompt across different seeds:
\textbf{the first column} (Original) is generated with default embeddings and consistently reproduces the same image regardless of the seeds. 
\textbf{The remaining five columns (Ours)} are generated with our proposed mitigation under five different seeds.
Additional results are provided in Appendix~\ref{app:more results on mitigation}.
}
    \label{fig:main}
\vspace{-0.3em}
\end{figure*}
Our empirical analysis reveals a surprising trend: Stable Diffusion does not rely heavily on the $\bt_i$ during generation. 
Instead, $\bp_i$ play an unexpectedly dominant role, particularly in memorized prompts.
We attribute this to a fundamental misalignment between how CLIP is trained and how diffusion models condition on input text embeddings.
Specifically, CLIP is trained via contrastive learning to optimize only $\bE$ to represent entire sentences, ignoring both $\bt_i$ and $\bp_i$.
Diffusion models, however, condition on all embeddings during generation, leaving them exposed to embeddings that are not explicitly optimized.

The issue is further exacerbated by the tokenizer’s design: 
prompts shorter than 77 tokens are padded by repeating \eot.
As a result, the diffusion model attends to multiple near-identical $\bE$ in the form of $\bp_i$, unintentionally amplifying their influence and increasing memorization risk.
Indeed, we find that all previously reported memorized prompts are fewer than 40 tokens,
resulting in over 30 near duplicated $\bE$ (see Appendix~\ref{app:setup}).

Consistent with this mechanism, we confirm that the vulnerability largely disappears in Stable Diffusion~v2.1.
While v2.1 adopts OpenCLIP~\citep{cherti2023reproducible} primarily for performance improvements, it also unintentionally removes \eot from padding positions.
By removing duplicated $\bE$, v2.1 substantially reduces exact-match memorization. This direct correspondence between our analysis and the behavior of v2.1 provides further support for our central claim.

Building on this insight, we propose two simple yet effective inference-time mitigation strategies that target the overemphasized role of $\bE$ and $\bp_i$: (1) a tokenizer-level fix that replaces the default \pad from \eot to the \texttt{!} token and masks the $\bE$; (2) partially masking the $\bp_i$. 
Both methods significantly reduce memorization without degrading image quality or increasing computational cost.
Crucially, our mitigation strategies not only suppress memorization but also restore diversity across different random seeds, 
preventing deterministic reproduction of memorized content 
and enabling 
richer generations even under identical prompts.
Unlike prior approaches that rely on dataset filtering, prompt tuning, or model retraining, 
our approach requires no model modifications and can be applied efficiently, making them highly practical for real-world deployment.
Our contributions can be summarized as follows:
\begin{itemize}
    \item We present a new perspective on memorization by shifting the focus from model internals to the overlooked structure of text embeddings. 
    We show that a mismatch between training dynamics of CLIP and its use during diffusion model inference introduces structural biases.
    \item We discover that $\bp_i$ are structural duplications of $\bE$, unintentionally amplifying its influence during generation. This embedding-level mechanism creates a new pathway for memorization distinct from data duplication.
    \item We propose two simple inference-time mitigation strategies: 
    (1) replacing the tokenizer’s default \pad (\eot) with a neutral token (\texttt{!}) and masking the single $\bE$; 
    (2) partially masking $\bp_i$.
    Both approaches suppress memorization effectively while preserving image quality and prompt alignment.
\end{itemize}

\begin{figure}[h]
    \centering
    \includegraphics[width=0.5\textwidth]{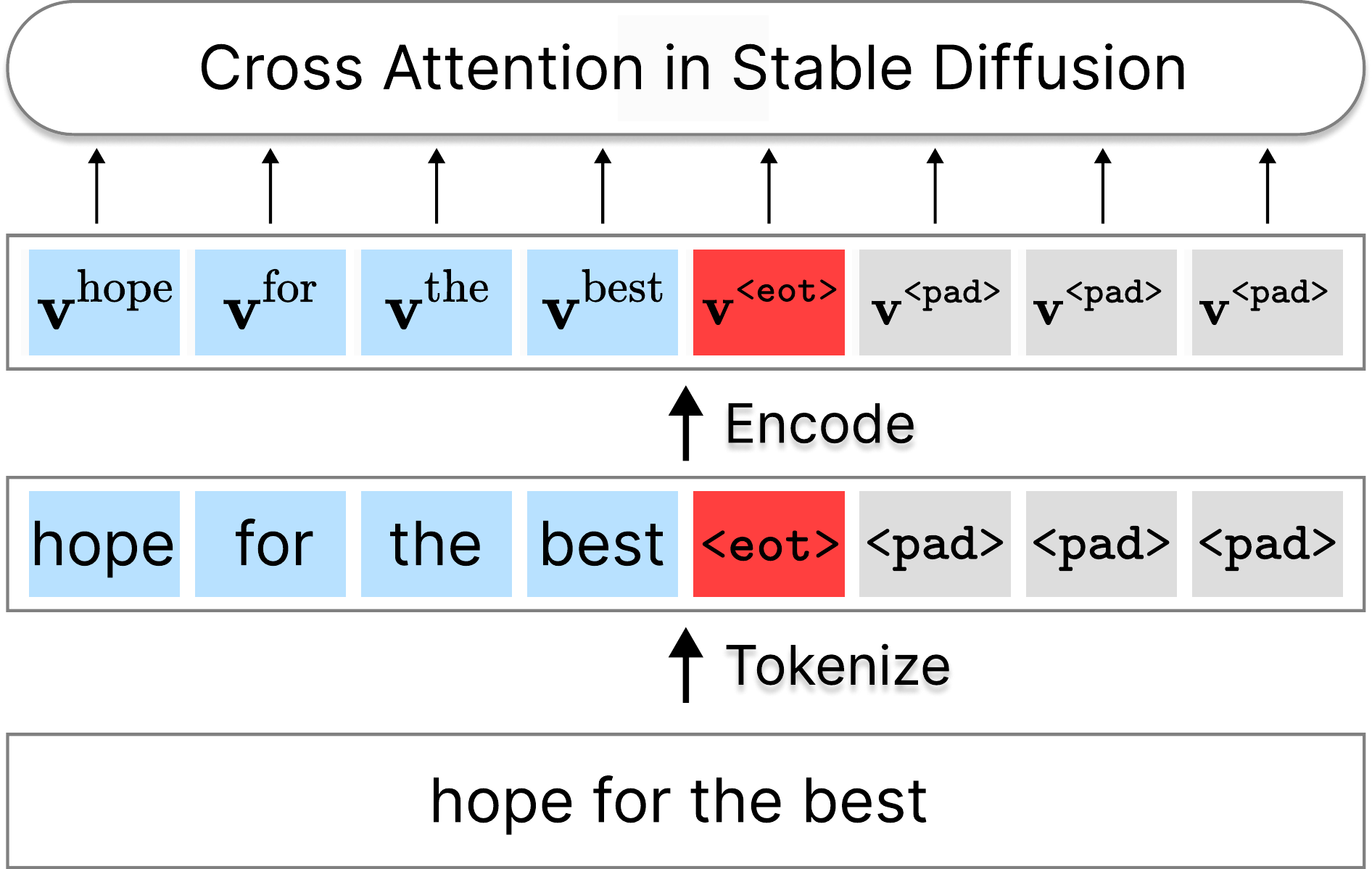}
    \caption{
    A text prompt is first processed by the tokenizer into discrete tokens,
    which are then transformed by the CLIPText into embeddings.
    All embeddings, including $\bt_i$ and $\bp_i$, serve as conditioning inputs for the model.
    }
    \label{fig:diff}
    \vspace{-0.5em}
\end{figure}

\section{Preliminaries}
\label{preliminaries}
\subsection{CLIP Embeddings and Training Dynamics}
\label{sec:clip}
CLIP~\citep{radford2021learning} is a multimodal model that embeds both images and text into a shared representation space.
It consists of a Text Encoder and an Image Encoder,
whose embeddings are optimized through contrastive learning.
This process maximizes the similarity of text-image pairs while pushing apart unpaired embeddings.
A key aspect of CLIP’s text embedding module is its reliance on the $\bE$ as the main representative of textual information.
CLIP focuses on learning a compact representation centered around $\bE$.
While all tokens in the text sequence contribute to the final embedding, their role is largely implicit,
as the optimization during training predominantly emphasizes the representation quality of $\bE$.
Similarly, CLIP’s Image Encoder processes an input image through a Vision Transformer, where each patch is embedded and then aggregated via a learnable \texttt{[CLS]}.  
The \texttt{[CLS]} embedding serves as a summary representation of the entire image.  
This design choice simplifies downstream applications and ensures alignment between textual and visual modalities 
in the shared embedding space.
The dominance of $\bE$ in CLIP’s text embedding process provides theoretical support for 
prior findings \citep{yi2024towards,chenexploring} that highlight the importance of \eot.

\subsection{T2I Diffusion Model and Stable Diffusion}
\label{sec:pre diffusion}
Text-to-image diffusion models, such as Stable Diffusion~\citep{rombach2022high, podellsdxl}, 
leverage pretrained multimodal representations to generate images from prompts.
A common text encoder in diffusion models is CLIPText,
which transforms input prompts into a structured embedding representation.
While CLIPText training emphasizes $\bE$ as the primary representation, Stable Diffusion incorporates all token embeddings,
including those only implicitly trained through contrastive learning.
This distinction matters, as Stable Diffusion distributes attention across the entire sequence rather than relying on a single representative embedding.

Specifically, the input prompt is tokenized and embedded into a sequence with a maximum length of 77.
Prompts exceeding this limit are truncated, while shorter prompts are padded for uniform length.
These embeddings are processed in the cross-attention layers,
allowing the diffusion model to condition on the entire embedding sequence during generation as illustrated in Fig.~\ref{fig:diff}.
This design introduces potential side effects, as the model may attend to embeddings
that were not explicitly trained to carry semantic meaning, 
potentially resulting in unintended behavior.

\subsection{Memorization in Diffusion}
Since the discovery of memorization~\citep{carlini2023extracting} in Stable Diffusion, numerous studies have sought to understand its underlying causes and mechanisms from diverse perspectives.
\citet{somepalli2023understanding} showed that caption specificity 
is a major driver of memorization, and proposed multi-caption training and randomized token perturbations for mitigation. 
\citet{gumemorization} pointed out that small dataset size and uninformative conditioning signals are critical factors behind memorization.
\citet{wen2024detecting} detected memorization by analyzing the magnitude of text-conditional noise predictions, identified trigger tokens, and mitigated memorization by perturbing tokens.
\citet{jeon2025understanding} adopted a geometric framework to demonstrate that memorization corresponds to sharp, isolated peaks in the learned density and mitigated it by optimizing the initial Gaussian noise to lie in smoother regions.
Several works have analyzed cross-attention patterns to explain memorization.  
\citet{chenexploring} discovered the ``Bright Ending'' phenomenon, where attention concentrates on the \eot during the final denoising steps,  
while \citet{ren2024unveiling} observed that memorized prompts exhibit lower cross-attention entropy, with attention disproportionately concentrated on specific trigger tokens. 

\section{Rethinking the Importance of Embeddings}
\label{sec:re-evaluating}
Text-to-Image Diffusion models are conditioned on a sequence of text embeddings derived from user prompts.
As discussed in Section~\ref{sec:pre diffusion}, there is a fundamental discrepancy between CLIP’s training objective, which optimizes only the $\bE$ through contrastive learning, and the inference of models, which rely on the full embedding sequence including $\bt_i$ and $\bp_i$.
This misalignment suggests that the factors influencing memorization may arise from structural properties of the embedding space rather than tokens.
Prior studies~\citep{somepalli2023understanding, wen2024detecting} have focused on ``trigger tokens,'' showing that certain tokens strongly drive memorization.
These findings are valid at the token level: modifying a trigger token before embedding can alter the output.

However, once the token sequence is encoded by CLIP, its causal encoder aggregates information across the prompt.
As a result, the influence that trigger tokens exert at the token level is largely absorbed into $\bE$, and the individual $\bt_i$ play a substantially reduced role.
This illustrates a key point: the importance of a token before embedding does not translate directly to its importance after embedding.
Because diffusion models condition on all embeddings, not the raw tokens, understanding memorization requires analyzing how each embedding contributes to the generation.
Motivated by this gap, we shift the analysis from token-level to embedding-level mechanisms.
We perform a series of controlled embedding interventions to quantify how each element of the CLIP embedding sequence contributes to memorization.
Surprisingly, we find that the $\bt_i$ contribute minimally to memorization,
while $\bp_i$, which are often dismissed as mere placeholders, 
exert a much stronger influence in memorized prompts.

\begin{mdframed}[backgroundcolor=gray!10, linewidth=0.5pt, roundcorner=5pt]
\textbf{Claim 1}: Prompt embeddings ($\bt_i$) have surprisingly minimal influence on memorization. \\
\textbf{Claim 2}: Padding embeddings ($\bp_i$) play a dominant role in both memorization and image quality, contradicting prior assumptions of their irrelevance.
\end{mdframed}

\paragraph{Model.}
To ensure a fair comparison with existing studies, we focus our experiments on Stable Diffusion~v1.4, which has been the standard in most prior memorization research~\citep{wen2024detecting,ren2024unveiling,chenexploring,hintersdorf2024finding,hongmembench,jeon2025understanding,jiangimage,kowalczuk2025finding,kim2025diffusion} and provides a well-established baseline. 
In this context, Stable Diffusion~v1.4 naturally serves as the most appropriate choice for studying memorization.
We therefore treat Stable Diffusion~v1.4 as the canonical benchmark for memorization.
Although newer models such as Stable Diffusion~v3~\citep{esser2024scaling} and FLUX~\citep{flux2024} have emerged, they have not yet been systematically examined for memorization, and thus currently lack the context and benchmarks required for meaningful comparison.
\paragraph{Dataset.}
\citet{webster2023reproducible} released 500 candidate memorized prompts and labeled 345 of them as Matching Verbatim (MV), Retrieval Verbatim (RV), or Template Verbatim (TV). Prior studies~\citep{wen2024detecting,ren2024unveiling,chenexploring,kowalczuk2025finding} typically used all 500 prompts.
However, not all 500 qualify as reliable “memorized” cases: some reproduce only under specific random seeds, some yield severely degraded images, and others appear to match training images that are not the intended targets. Such issues can contaminate subsequent analyses. Examples of problematic cases are provided in Appendix~\ref{app:dataset analysis}.

To avoid this, we restrict our primary analysis to MV prompts, which reproduce a near-identical copy of a training image consistently across seeds.
However, the original set includes only 86 MV prompts, limiting statistical reliability.
To expand this set, we leverage a subset of Membench~\citep{hongmembench} and apply two strict filtering criteria: \textbf{(1) Exact Match:} The prompt must reproduce a single training image. \textbf{(2) Consistency:} For each prompt, images generated using 10 different seeds must exhibit an SSCD similarity score~\citep{pizzi2022self} $\geq 0.5$ to the training image.
This yields 372 additional prompts, resulting in a final set of 458 prompts.
To assess the generalizability and robustness beyond memorized cases, we also evaluate 500 non-memorized prompts each from LAION~\citep{schuhmann2022laion}, COCO~\citep{lin2014microsoft}, and Lexica~\citep{santana_gustavostastable-diffusion-prompts_2022}. Results on non-memorized prompts are reported in Appendix~\ref{app:non}.

\paragraph{Evaluation Metrics.}
Following prior work on memorization in Stable Diffusion~\citep{somepalli2023understanding,chenexploring,wen2024detecting,kowalczuk2025finding,jiangimage}, we adopt SSCD~\citep{pizzi2022self} as the primary metric to determine whether memorization occurs. 
To ensure a stricter and more conservative evaluation, we follow the most rigorous threshold used in previous studies, setting SSCD~$\geq 0.5$ as the criterion for memorization.
Instead of comparing to the real training image, we measure the SSCD between each intervention output and the image generated using the original embedding. A score near 0.5 indicates minimal visual difference from the image generated using the original embedding.
For generation quality, we use CLIPScore~\citep{hessel2021clipscore} to assess text-image alignment, and Aesthetic Score~\citep{schuhmann2022laion} to evaluate overall visual quality, as it provides a more direct and interpretable  
measure of per-image quality than FID~\citep{hongmembench}.
Results are averaged over 10 random seeds; means are computed over all generated images, and $\pm$ values denote the average image-wise standard deviation across seeds.

\subsection{Prompt Embeddings Play a Minor Role}
As discussed above, $\bt_i$ are not explicitly optimized during CLIP training. 
Yet Stable Diffusion conditions on the full embedding sequence, 
raising the question of how much influence they actually have during generation. 
To probe this, we conduct a series of interventions targeting $\bt_i$:
\begin{flalign*}
& \textit{(a)} \; \textbf{$\bE$ and $\bp_i$ are masked:} & \\ 
& \qquad \mathbf{Emb} = [\bS,\bt_1,\dots,\bt_n,\mathbf{0},\dots,\mathbf{0}] & \\
& \textit{(b)} \; \textbf{$\bt_i$ are replaced with $\bE$:} & \\
& \qquad \mathbf{Emb} = [\bS,\bE,\dots,\bE,\bp_1,\dots,\bp_d] & \\
& \textit{(c)} \; \textbf{$\bt_i$ are masked:} & \\
& \qquad \mathbf{Emb} = [\bS,\mathbf{0},\dots,\mathbf{0},\bE,\bp_1,\dots,\bp_d] &
\end{flalign*}
In \textit{(a)}, we retain only $\bt_i$, masking both $\bE$ and $\bp_i$ with zero vectors.
If $\bt_i$ carries meaningful information, the output should remain visually plausible.
However, the generated images exhibit severe degradation, as evidenced by sharp drops in both CLIPScore and Aesthetic Score in Table~\ref{tab:prompt intervention}, demonstrating that $\bt_i$ alone is insufficient to drive the generation process.
\begin{figure}[h]
    \centering
    \includegraphics[width=0.5\textwidth]{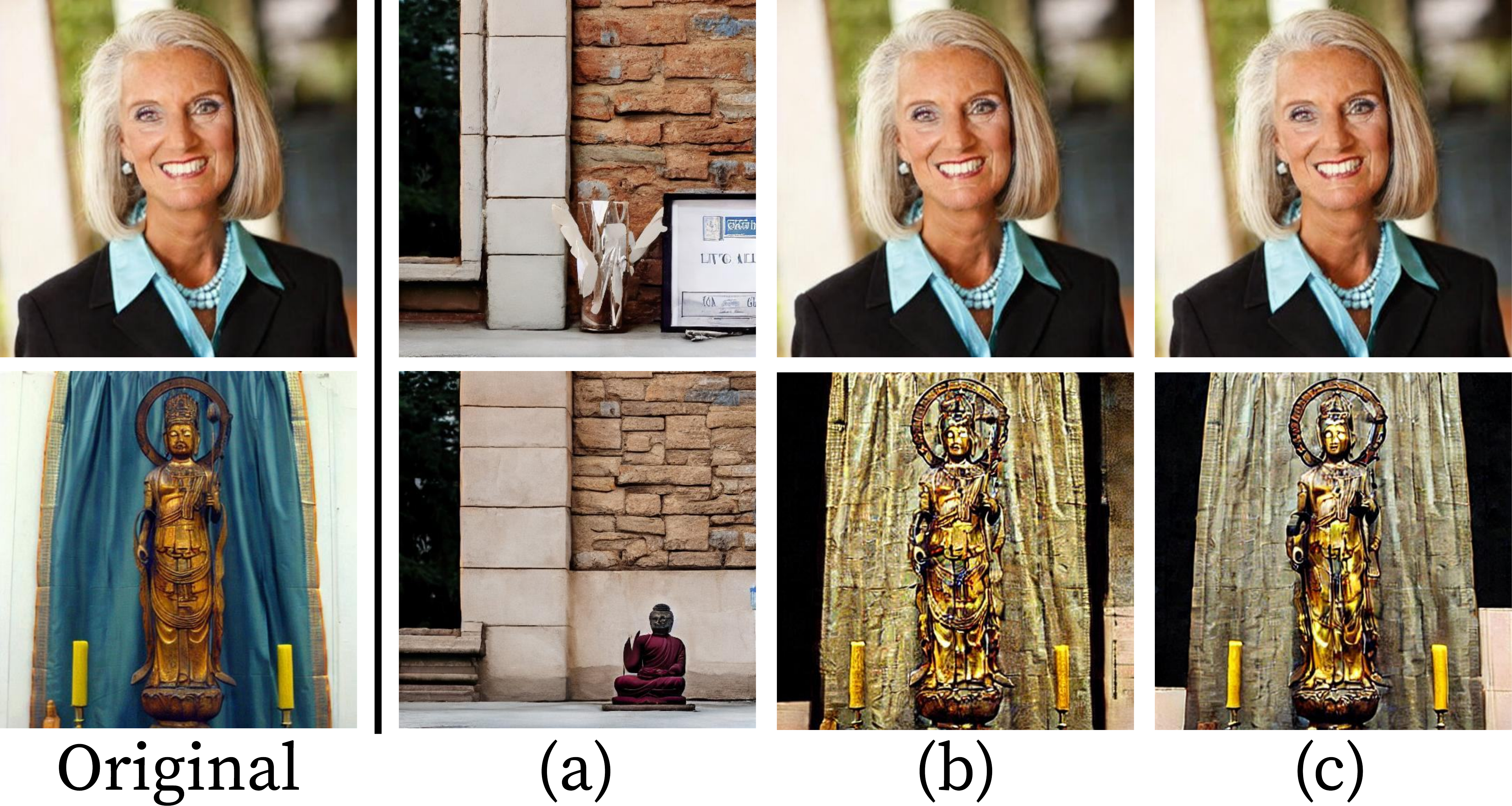}
\caption{\textbf{$\bt_i$ are not important.}
(a) shows that $\bt_i$ alone cannot guide generation, as the resulting images collapse when they are the only embeddings retained.
In contrast, (b) replacing $\bt_i$ with $\bE$ or (c) masking $\bt_i$ preserve most of the original structure.
}
\label{fig:prompt intervention}
\vspace{-0.5em}
\end{figure}
In contrast, \textit{(b)} replacing all $\bt_i$ with $\bE$ 
and \textit{(c)} masking $\bt_i$ yield outputs that are visually similar to the original generation. 
As shown in Fig.~\ref{fig:prompt intervention}, despite slight variations, the outputs maintain overall structure and semantic alignment. 
This is further supported by SSCD scores near 0.5 (Table~\ref{tab:prompt intervention}), which reflect high similarity to the original images.
These findings challenge the common assumption that $\bt_i$ are the primary driver of image generation. Instead, they suggest that Stable Diffusion relies heavily on $\bE$ and $\bp_i$ for conditioning.
These results reinforce our earlier observation that token-level influence does not persist after CLIP embedding.
\begin{table}[h]
\centering
\begin{tabular}{c|c|c|c}
\toprule
\textbf{Method} & 
\textbf{SSCD} & \textbf{CLIPScore} & \textbf{Aesthetic} \\
\midrule
\textit{(a)}
    & 0.04 $\pm$ 0.05
    & 0.25 $\pm$ 0.06 
    & 5.02 $\pm$ 0.45 \\
\textit{(b)}
    & 0.55 $\pm$ 0.14 
    & 0.29 $\pm$ 0.02
    & 5.10 $\pm$ 0.23 \\
\textit{(c)}
    & 0.42 $\pm$ 0.18 
    & 0.29 $\pm$ 0.03
    & 5.15 $\pm$ 0.30 \\
\bottomrule
\end{tabular}
\caption{
The role of $\bt_i$ in memorized image generation.
\textit{(a)}~$\bE, \bp_i \rightarrow \bz$: only $\bt_i$ remain while $\bE$ and $\bp_i$ are masked.
\textit{(b)}~$\bt_i \rightarrow \bE$: $\bt_i$ are replaced with $\bE$.
\textit{(c)}~$\bt_i \rightarrow \bz$: $\bt_i$ are masked.}
\label{tab:prompt intervention}
\vspace{-0.8em}
\end{table}

\subsection{Padding Embeddings Play a Major Role}
\label{sec:pad is important}
Given the limited role of $\bt_i$, 
we next examine the contribution of the remaining embeddings. 
In the following experiments, we manipulate $\bE$ and $\bp_i$ in isolation or jointly, and measure their impact on memorized prompts.
We describe five intervention setups used in our analysis:
\begin{flalign*}
& \textit{(d)} \; \textbf{$\bp_i$ are replaced with $\bE$:} & \\
& \qquad \mathbf{Emb} = [\bS,\bt_1,\dots,\bt_n,\bE,\bE,\dots,\bE] & \\
& \textit{(e)} \; \textbf{$\bt_i$, $\bp_i$ are replaced with $\bE$:} & \\
& \qquad \mathbf{Emb} = [\bS,\bE,\dots,\bE] & \\
& \textit{(f)} \; \textbf{$\bE$ is masked:} & \\
& \qquad \mathbf{Emb} = [\bS,\bt_1,\dots,\bt_n,\bz,\bp_1,\dots,\bp_d] & \\
& \textit{(g)} \; \textbf{$\bt_i$, $\bE$ are replaced with $\bp_{\text{mean}}$:} & \\
& \qquad \mathbf{Emb} = [\bS,\bp_{\text{mean}},\dots,\bp_{\text{mean}}, \bp_1, \dots, \bp_d] & \\
& \textit{(h)} \; \textbf{$\bp_i$ are masked:} & \\
& \qquad \mathbf{Emb} = [\bS,\bt_1,\dots,\bt_n,\bE,\bz,\dots,\bz] &
\end{flalign*}

The importance of the $\bE$ in generation and memorization has been highlighted in prior works. \citet{yi2024towards} observed that the generation process is largely determined by $\bE$ at earlier stages of denoising,
and \citet{chenexploring} identified the ``Bright Ending'',
where $\bE$ attracts unusually high attention in memorized prompts.
In \textit{(d)}, we replace all $\bp_i$ with $\bE$, and in \textit{(e)} we replace both $\bt_i$ and $\bp_i$ with $\bE$, retaining only $\bE$.
In both cases, the generated outputs remain very close to the original images, preserving overall structure and yielding SSCD values near 0.5 (Table~\ref{tab:pad intervention}). 
These results reinforce the established importance of $\bE$ in guiding generation. 
However, when $\bE$ is masked
in \textit{(f)}, the results remain virtually identical to the original. This suggests that generation is not solely dependent on $\bE$.
\begin{figure}[h]
    \centering
    \includegraphics[width=0.5\textwidth]{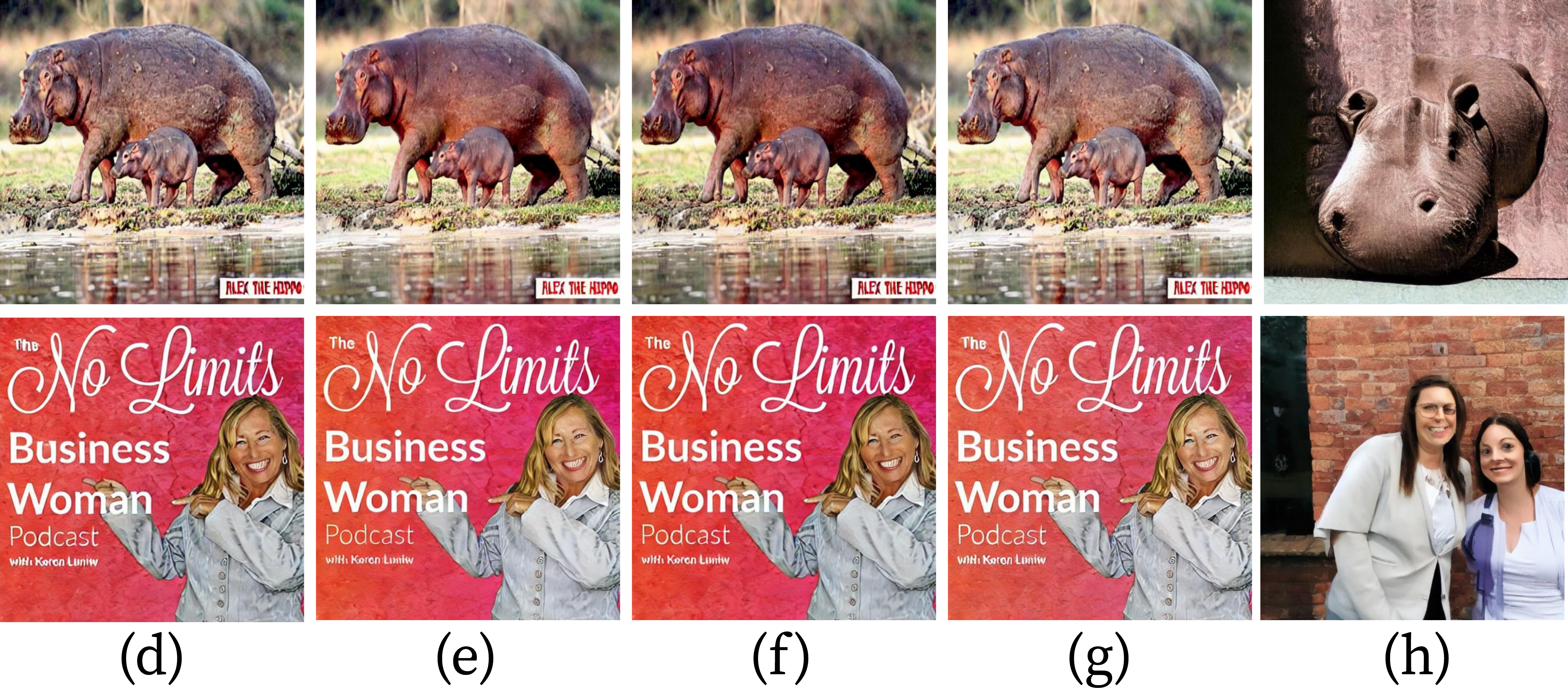}
\caption{\textbf{$\bp_i$ are important.}
(d) to (g) show cases where $\bp_i$ are preserved, either by replacing $\bp_i$ with $\bE$, 
remaining only $\bE$, masking $\bE$, or remaining only $\bp_i$. 
All four preserve the structure and semantics of the original image, whereas (h) masking $\bp_i$ generates images that deviate significantly from it.}
    \label{fig:prompt embedding}
    \vspace{-1.2em}
\end{figure}
\begin{table}[h]
\centering
\begin{tabular}{c|c|c|c}
\toprule
\textbf{Method} & \textbf{SSCD} & \textbf{CLIPScore} & \textbf{Aesthetic} \\
\midrule
\textit{(d)} 
    & 0.85 $\pm$ 0.07 & 0.32 $\pm$ 0.01 & 5.28 $\pm$ 0.07 \\
\textit{(e)} 
    & 0.49 $\pm$ 0.13 & 0.29 $\pm$ 0.02 & 5.05 $\pm$ 0.24 \\
\textit{(f)} 
    & 0.95 $\pm$ 0.05 & 0.32 $\pm$ 0.01 & 5.31 $\pm$ 0.07 \\
\textit{(g)} 
    & 0.50 $\pm$ 0.17 & 0.29 $\pm$ 0.02 & 5.09 $\pm$ 0.28 \\
\textit{(h)} 
    & 0.07  $\pm$ 0.06 & 0.29 $\pm$ 0.04 & 5.10 $\pm$ 0.42 \\
\bottomrule
\end{tabular}
\caption{
The role of $\bE$ and $\bp_i$ in memorized image generation.
\textit{(d)}~$\bp_i \rightarrow \bE$: $\bp_i$ are replaced with $\bE$. 
\textit{(e)}~$\bt_i, \bp_i \rightarrow \bE$: $\bt_i$ and $\bp_i$ are replaced with $\bE$. 
\textit{(f)}~$\bE \rightarrow \bz$: $\bE$ is masked.
\textit{(g)}~$\bt_i, \bE \rightarrow \bp_{\text{mean}}$: $\bt_i$ and $\bE$ are replaced with the mean of $\bp_i$. 
\textit{(h)}~$\bp_i \rightarrow \bz$: $\bp_i$ are masked, resulting in collapse.
}
\label{tab:pad intervention}
\vspace{-0.8em}
\end{table}

To further explore this, 
we test whether $\bp_i$ alone can preserve generation quality. 
In \textit{(g)}, we replace both $\bt_i$ and $\bE$ with $\bp_i$ (using $\bp_{\text{mean}}$ to compensate for the insufficient length). 
Surprisingly, the model still produces well-structured and semantically coherent images.
This result suggests that $\bp_i$ can functionally substitute for $\bE$ in memorized prompts.
Finally, in \textit{(h)}, we isolate the effect of $\bp_i$ by masking them. 
This intervention leads to a significant collapse in generation, 
with SSCD dropping to near zero,
indicating the generated image is entirely different from the original.
\begin{figure*}[h]
    \centering
    \includegraphics[width=0.95\textwidth]{figure/3_2/cross_attn_2.pdf}
    \caption{
    While \citet{chenexploring} reported a “Bright Ending”, an attention spike on \eot (specifically on $\bE$) when memorization occurs, we observe that this effect is not limited to $\bE$. Multiple \pad ($\bp$) adjacent to \eot also exhibit elevated attention, 
    indicating that they play a substantial role in the memorization process rather than serving as placeholders.}
    \label{fig:cross_attn}
    \vspace{-1em}
\end{figure*}
These findings challenge the common assumption that $\bp_i$ are semantically irrelevant, reinforced by \citet{toker2025padding}, which argued that $\bp$ are unlikely to carry meaningful information in models with frozen text encoders such as Stable Diffusion.
Contrary to this belief, our experiments reveal that $\bp_i$ play an active role in generation and are critical for reproducing memorized content, functioning nearly as strongly as $\bE$.
Moreover, while \citet{chenexploring} reported the “Bright Ending”, where cross-attention excessively concentrates on $\bE$ during the final denoising steps, we find that this effect extends beyond $\bE$. 
As shown in Fig.~\ref{fig:cross_attn}, multiple $\bp$ adjacent to $\bE$ also exhibit elevated attention, indicating that $\bE$ and its neighboring $\bp$ jointly amplify the signal. 
This observation further supports our claim that both $\bE$ and $\bp$ exert substantial influence on memorization.

\section{Understanding~\&~Mitigating~Memorization}
\subsection{Repeated \eot Drives Memorization}
\label{sec:pad into !}
\begin{figure*}[h]
    \centering
    \includegraphics[width=0.73\textwidth]{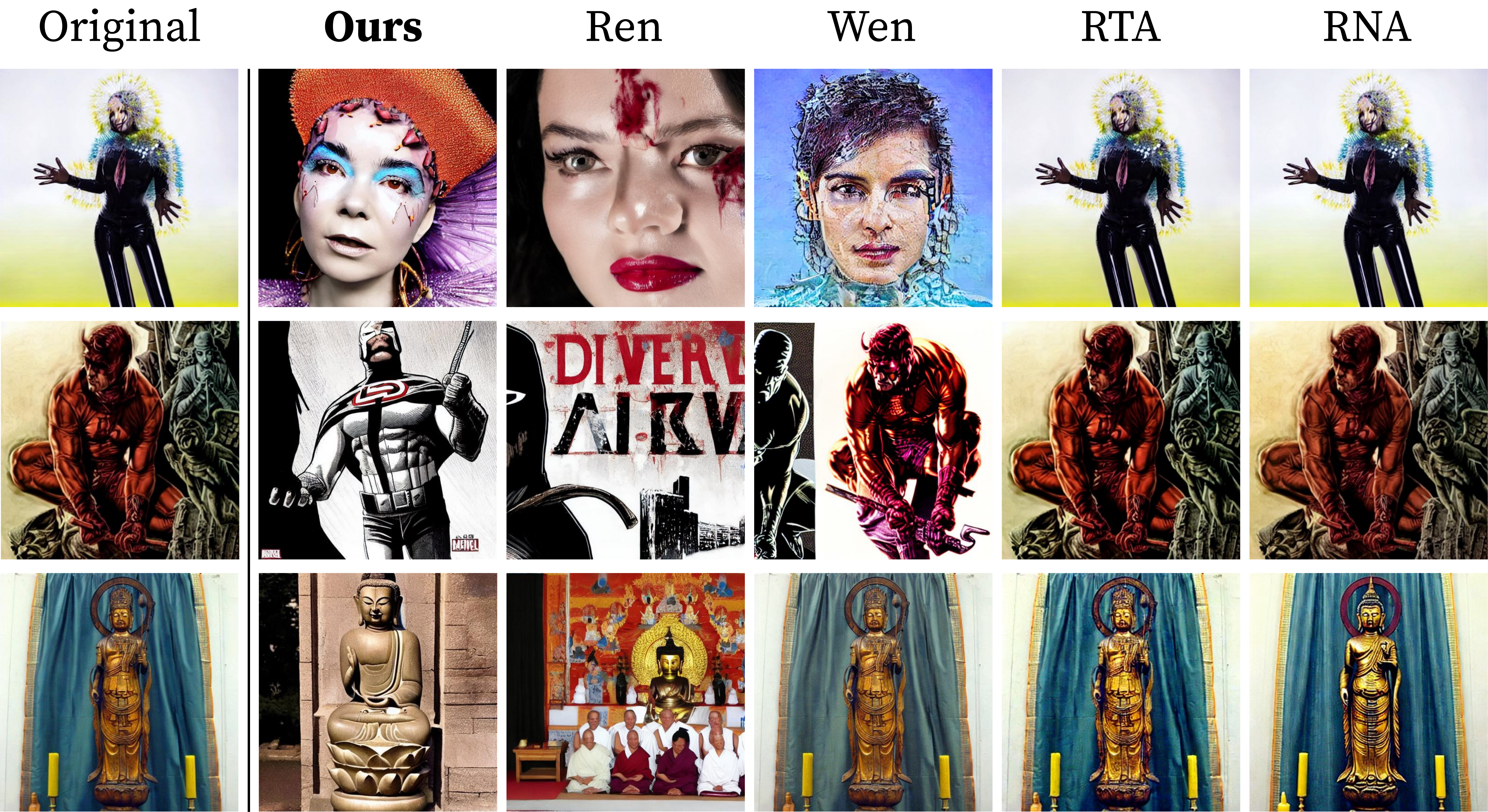}
    \caption{\textbf{Comparison of mitigation results across five methods using the same prompt and seed.} Our method (second column) mitigates memorization while preserving structure and prompt alignment.
\citet{ren2024unveiling} reduce memorization but introduce distortion, while \citet{wen2024detecting}, RTA and RNA~\citep{somepalli2023understanding} often fail to mitigate memorization. Additional examples are in Appendix~\ref{app:comparison}.}
    \label{fig:comparison}
\end{figure*}

Our experiments reveal that the $\bp_i$ play a disproportionately large role in memorization.
We hypothesize that this effect arises from a structural flaw in the tokenizer design of Stable Diffusion~v1.4.
As detailed in Section~\ref{sec:pre diffusion}, this version uses CLIPText, which pads all prompts to a fixed length of 77 tokens by repeatedly inserting the \eot.

This results in multiple occurrences of the same token, \eot, within a single input sequence: once at its intended position and repeatedly in the \pad region.
Because these tokens are identical, the resulting embeddings are also nearly identical (see Appendix~\ref{app:clip and openclip} for PCA and t-SNE visualizations).
Moreover, CLIP is trained via contrastive learning where only $\bE$ is explicitly optimized to represent sentence-level semantics.
This makes $\bE$ far more semantically meaningful than other embeddings.

Our key claim is that the repetition of $\bE$ across the input sequence unintentionally amplifies its effect during conditioning in diffusion,
leading the model to overfit to specific examples, especially for short prompts where the padding region occupies the majority of the input.

To further validate this claim, we compare Stable Diffusion~v1.4 and v2.1.
We observe that exact matching memorization is significantly less frequent in v2.1.
While this reduction has often been attributed to dataset de-duplication~\citep{webster2023reproducible},
prior work~\citep{somepalli2023understanding} argued that data de-duplication alone cannot fully eliminate memorization.
We argue that this phenomenon is consistent with the removal of \eot\ from the padding region.
Because \eot no longer appears in padded positions, its influence on the model’s attention and embedding dynamics is naturally diminished, leading to a substantial drop in memorization.

Notably, v2.1 replaces the original CLIP text encoder with OpenCLIP, primarily for performance improvements.
However, to the best of our knowledge, no official documentation or prior work on v2.1 has discussed the motivation behind replacing \eot with a semantically neutral padding token.
This modification was likely incidental rather than intentional,
yet it inadvertently removed a key source of memorization observed in v1.4.
Thus, the absence of \eot duplication, rather than dataset de-duplication or architectural changes, serves as the major driver
explaining why v2.1 exhibits almost no exact matching memorization.

Consequently, this design change reduces the model’s reliance on the semantically dominant embedding $\bE$, thereby lowering the risk of memorization during both training and inference.
Taken together, these findings support our central claim: the unintended duplication of $\bE$ in the padding region is a key mechanism underlying memorization in Stable Diffusion.
\begin{table*}[h]
    \centering
    \begin{tabular}{l|l|c|c|c|c}
    \toprule
    \textbf{Category} & \textbf{Method} & \textbf{SSCD} & \textbf{CLIPScore} & \textbf{Aesthetic} & \textbf{LPIPS} \\
    \midrule
    \multirow{2}{*}{Replace \pad Token (Ours)} 
    & \pad $\rightarrow$ \texttt{!} \& $\bE \rightarrow \bz$ & 0.08 $\pm$ 0.06 & 0.31 $\pm$ 0.03 & 5.09 $\pm$ 0.44 & 0.67 $ \pm$ 0.05 \\
    & \pad $\rightarrow$ \texttt{!}  & 0.14 $\pm$ 0.08 & 0.32 $\pm$ 0.03 & 5.21 $\pm$ 0.38 & 0.64 $\pm$ 0.09 \\
    \multirow{1}{*}{Mask $\bp_i$ Embedding (Ours)} 
    & $\bp_i \rightarrow \bz$ (70\%) & 0.10 $\pm$ 0.07 & 0.31 $\pm$ 0.04 & 5.13 $\pm$ 0.41 & 0.66 $\pm$ 0.06 \\
    \midrule
    \multirow{4}{*}{Prior Works}
    & \citet{ren2024unveiling} & 0.09 $\pm$ 0.06 & 0.30 $\pm$ 0.03 & 5.18 $\pm$ 0.39 & 0.65 $\pm$ 0.05\\
    & \citet{wen2024detecting} & 0.51 $\pm$ 0.35 & 0.32 $\pm$ 0.03 & 5.22 $\pm$ 0.33 & 0.49 $\pm$ 0.19 \\
    & RTA~\citep{somepalli2023understanding} & 0.58 $\pm$ 0.32 & 0.31 $\pm$ 0.03 & 5.23 $\pm$ 0.31 & 0.42 $\pm$ 0.22 \\
    & RNA~\citep{somepalli2023understanding} & 0.53 $\pm$ 0.28 & 0.31 $\pm$ 0.03 & 5.20 $\pm$ 0.32 & 0.44 $\pm$ 0.23 \\
    \bottomrule
    \end{tabular}
    \caption{
    Simply changing the \pad token mitigates a substantial portion of memorization, 
    while additionally masking $\bE$ further suppresses leakage. 
    We also introduce partial masking of $\bp_i$ as a flexible alternative that enables fine-grained control.
    Compared to prior works (bottom), our method achieves low SSCD while maintaining prompt alignment, image quality, and perceptual diversity.}
    \label{tab:mitigation_comparison}
    \vspace{-0.5em}
\end{table*}

\subsection{Reducing \eot Overemphasis for Mitigation}
\paragraph{Decoupling $\bp_i$ from $\bE$.}
To mitigate memorization, we find that reducing the influence of $\bE$ is essential. We therefore modify the tokenizer to replace the default \pad token from \eot to a semantically neutral \texttt{!} token. This change ensures that \pad produce embeddings distinct from $\bE$, thus decoupling $\bp$ from $\bE$.
Empirically, this modification leads to a substantial reduction in memorization, as reflected in decreased SSCD scores (see Table~\ref{tab:mitigation_comparison}). 
To fully eliminate residual memorization, we further mask $\bE$ by replacing it with $\bz$.

Under this setting, memorization is almost completely removed (see Fig.~\ref{fig:main}).
These results confirm our hypothesis that the memorization risk is amplified by the repeated occurrence of $\bE$ in \pad positions. 
By structurally decoupling $\bp$ from $\bE$, we suppress memorization without degrading prompt alignment or image quality as evidenced by comparable CLIPScore and Aesthetic Score.

We compare our method with four prior inference-time mitigation strategies, with detailed descriptions provided in Appendix~\ref{app:comparison}.
As shown in Fig.~\ref{fig:comparison} and Table~\ref{tab:mitigation_comparison}, RNA, RTA~\citep{somepalli2023understanding}, and \citet{wen2024detecting} struggle to mitigate memorization, often reproducing the same memorized image across seeds.
\citet{ren2024unveiling} achieve competitive SSCD, but it degrades visual quality and semantic alignment.
In contrast, our approach not only achieves lower SSCD but also preserves image quality, 
maintaining alignment with the intended prompt. 
It is also the most efficient among all compared methods, 
with inference time nearly identical to the generation without mitigation (Appendix~\ref{app:comparison}). 
We also quantify cross-seed perceptual diversity using LPIPS~\citep{zhang2018unreasonable}. On memorized prompts, the original generations exhibit very low diversity (0.04), while ours significantly increases LPIPS, indicating restored sensitivity to random seeds.

\begin{figure*}[h]
    \centering
    \begin{minipage}{0.24\linewidth}
        \centering
        \includegraphics[width=\linewidth]{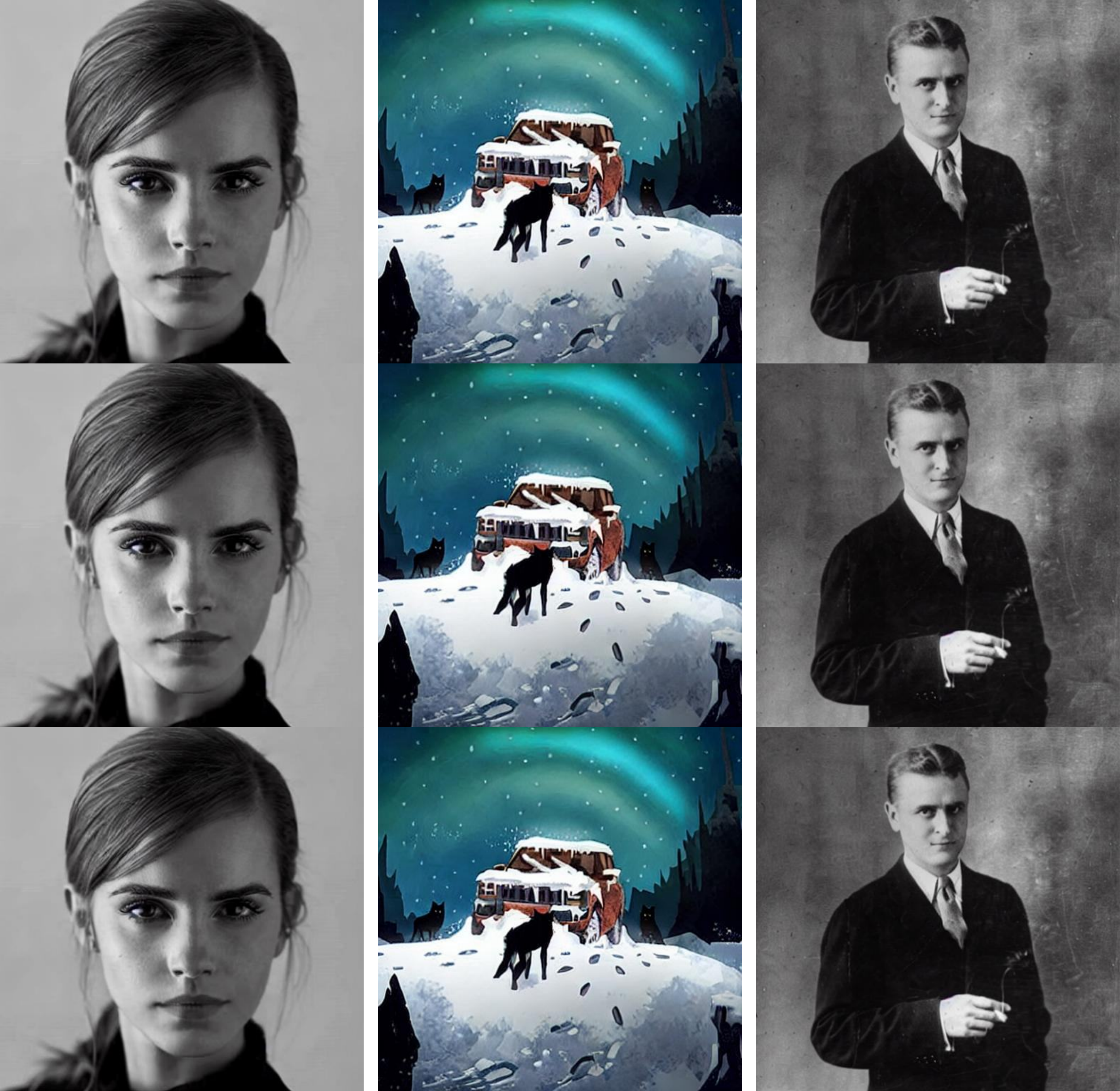}
        \textbf{(a)} Original
    \end{minipage}
    \hspace{2mm}
    \begin{minipage}{0.24\linewidth}
        \centering
        \includegraphics[width=\linewidth]{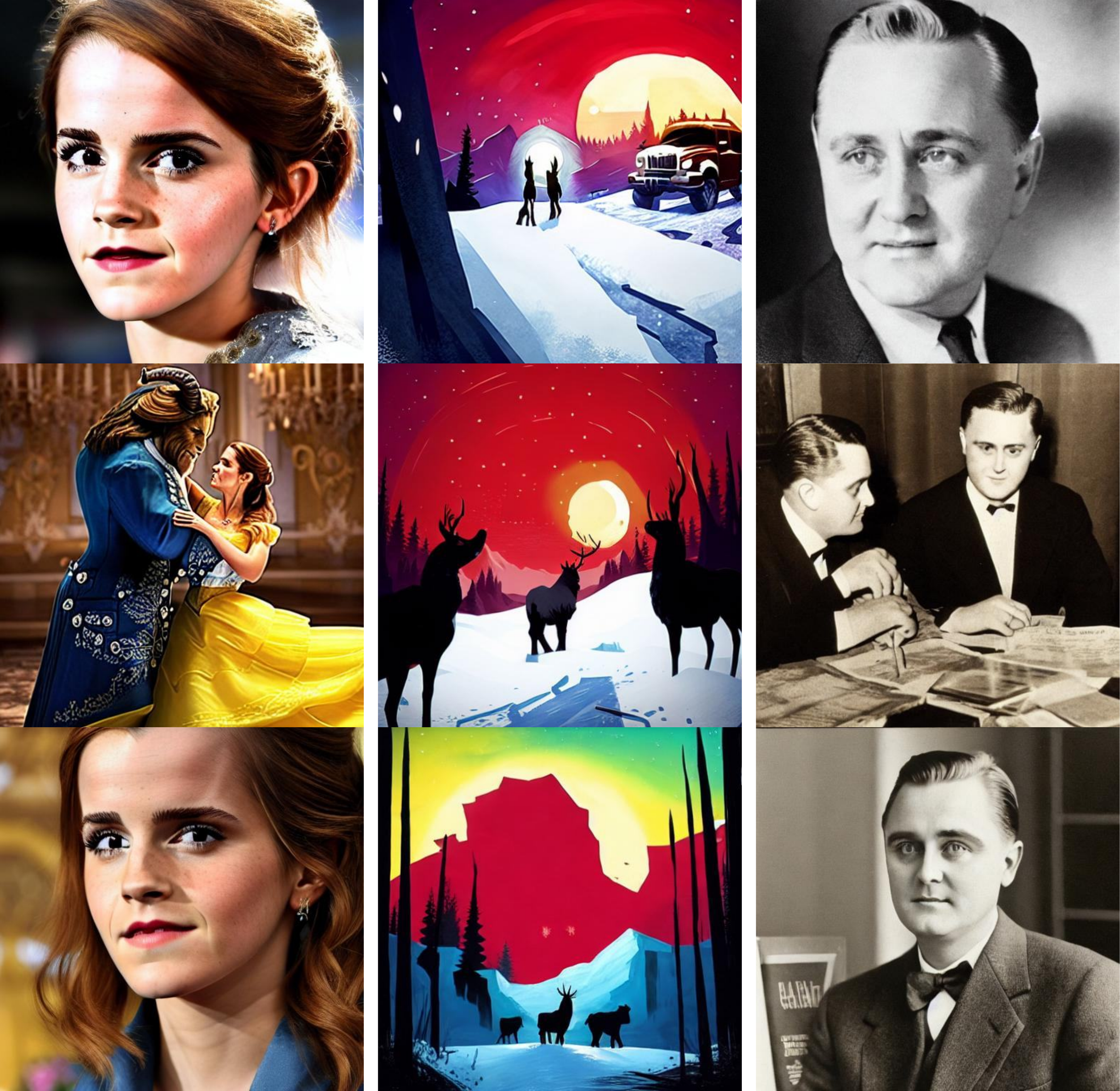}
        \textbf{(b)} Masking 70\%
    \end{minipage}
    \hspace{2mm}
    \begin{minipage}{0.24\linewidth}
        \centering
        \includegraphics[width=\linewidth]{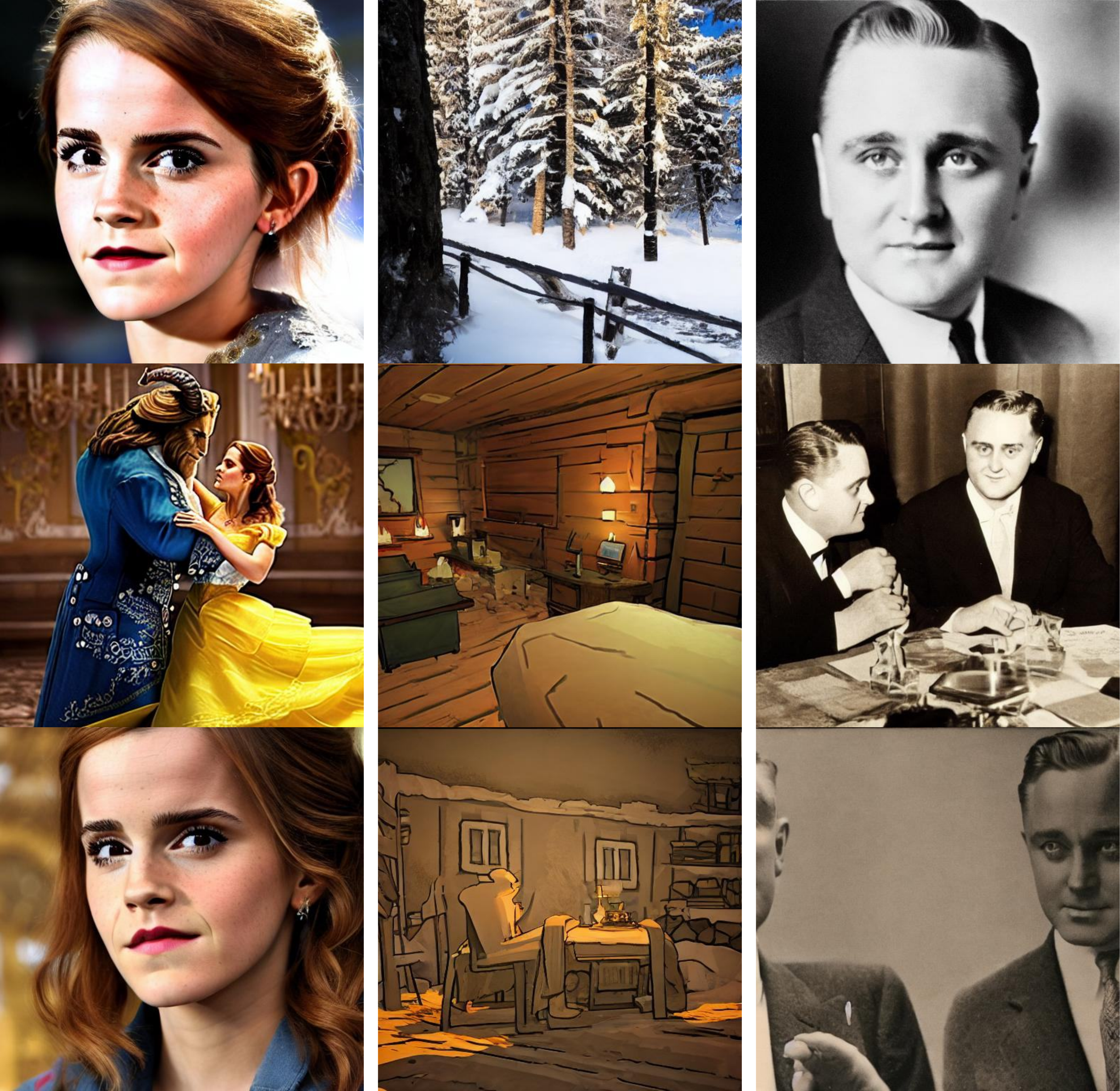}
        \textbf{(c)} Masking 100\%
    \end{minipage}
    \hspace{2mm}
    \caption{\textbf{Partial Masking $\mathbf{v}^{\text{pad}}_i$ to mitigate memorization.}  
    (b) and (c) show generations where 70\% and 100\% of the $\bp_i$ adjacent to $\bE$ are masked, respectively.
    Masking 70\% empirically strikes a well-balanced tradeoff, substantially reducing memorization with negligible quality degradation, whereas full masking (100\%) can often cause collapse or noticeable degradation.}
    \label{fig:partially masking}
    \vspace{-0.3em}
\end{figure*}

\paragraph{Partial Masking $\bp_i$.}
Building upon our previous findings, we propose an additional mitigation strategy grounded in the same core insight: reducing the influence of $\bE$ effectively suppresses memorization. 
While replacing the default \pad can reduce duplication at the tokenizer level (Section~\ref{sec:pad into !}), 
a complementary approach is to directly suppress the contribution of duplicated $\bE$ via masking.
A naive approach might suggest masking all $\bp_i$, but our experiments show that this often degrades generation quality. 
To balance suppression and stability, we apply \textit{partial masking} to the $\bp_i$.
This reduces the semantic redundancy induced by duplicated $\bE$, 
while preserving enough signal for the model to maintain generation fidelity.
Empirically, we find that masking 70\% of $\bp_i$ is sufficient to suppress memorization across most prompts while preserving semantic alignment and perceptual quality (Fig.~\ref{fig:partially masking}).
A key advantage of this method is its simplicity and flexibility: since it does not require tokenizer modification or model retraining, it can be easily integrated into existing pipelines.

\paragraph{Robustness and Effectiveness of Our Mitigation.}
Our method is simple, effective, and readily applicable in a plug-and-play manner. 
As shown in Table~\ref{tab:non-mem-eval}, applying our mitigation strategies to non-memorized prompts 
causes almost no degradation in image quality or prompt alignment compared to the original generation. 
Because it operates without requiring prior memorization detection, 
our approach eliminates additional detection overhead and can be seamlessly integrated into existing inference pipelines.
\begin{table}[h]
\centering
\begin{tabular}{l|c|c}
\toprule
\textbf{Method} & \textbf{CLIPScore} & \textbf{Aesthetic} \\
\midrule
Original & 0.35 $\pm$ 0.05 & 5.47 $\pm$ 0.51 \\
\midrule
\pad $\rightarrow$ \texttt{!} \& $\bE\rightarrow \bz$   & 0.34 $\pm$ 0.06 & 5.42 $\pm$ 0.51 \\
\pad $\rightarrow$ \texttt{!}  & 0.35 $\pm$ 0.05 & 5.49 $\pm$ 0.52 \\
$\bp_i \rightarrow \bz$  (70\%) & 0.34 $\pm$ 0.06 & 5.41 $\pm$ 0.51 \\
\bottomrule
\end{tabular}
\caption{Our methods preserve image quality: on 1,500 non-memorized prompts, CLIPScore and Aesthetic Score remain within the noise of the original, indicating no meaningful degradation or substantial deviation.}
\label{tab:non-mem-eval}
\vspace{-0.3em}
\end{table}

\begin{figure}[h]
    \centering
    \includegraphics[width=0.5\textwidth]{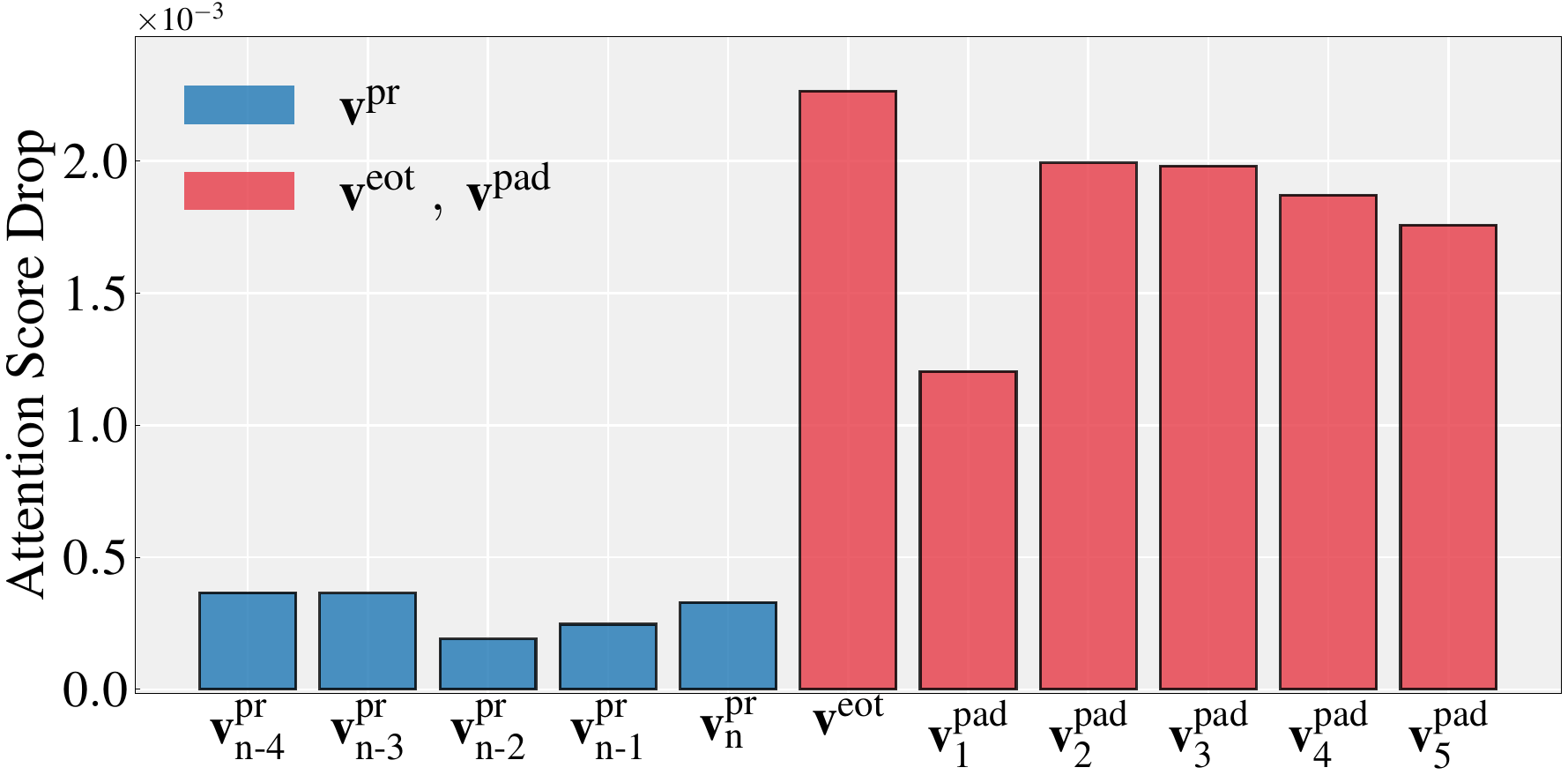}
    \caption{ 
    Using our mitigation, which replaces \pad with the \texttt{!} token and masks $\bE$, we observe that prompt region before $\bE$ remains largely unchanged, thereby preserving $\bt$ utility.
    The mitigation consistently suppresses attention on $\bE$ and $\bp$, effectively deactivating both embeddings. Overall, the mitigation removes the root cause without compromising the intended prompt semantics.
    }
    \label{fig:eot_centered_plots}
    \vspace{-1.5em}
\end{figure}

Furthermore, following the visualization method of \citet{chenexploring}, 
we re-examine token-wise attention activation after applying our mitigation. 
Since prompt lengths vary across samples, 
we align all sequences by the $\bE$ position and measure the average change in attention 
for the five $\bt$ preceding $\bE$ and the five $\bp$ following it. 
As shown in Fig.~\ref{fig:eot_centered_plots}, 
the attention drop for $\bt$ is negligible, 
indicating that the semantic content and utility of the prompt are preserved. 
In contrast, attention on $\bE$ and $\bp$ shows a substantial decrease, 
demonstrating that our mitigation effectively suppresses the components responsible for memorization 
while leaving the intended prompt semantics intact.

\section{Conclusion}
\label{sec:conclusion}
Stable Diffusion exhibits unexpected memorization due to a misalignment between CLIP's training objective and how diffusion models utilize embeddings.
Specifically, the repeated use of \eot as padding leads to multiple near-identical copies of the semantically dominant $\bE$, amplifying its influence.
We propose two simple mitigation strategies: (1) replacing \pad with a neutral token like \texttt{!} and masking $\bE$; (2) partially masking the $\bp_i$.
Both approaches reduce memorization without 
compromising image quality. 
Our findings highlight a structural cause of memorization and offer practical methods to address it.

\section*{Acknowledgements}
This work was supported in part by Institute of Information \& communications Technology Planning \& Evaluation (IITP) grant funded by the Korea government (MSIT) (No. RS-2024-00457882, AI Research Hub Project), IITP grant funded by the Korean Government (MSIT) (No. RS-2020- II201361, Artificial Intelligence Graduate School Program (Yonsei University)), and the National Research Foundation of Korea (NRF) grant funded by the Korea government (MSIT) (No. RS2025-23525649).

{
    \small
    \bibliographystyle{ieeenat_fullname}
    \bibliography{main}
}


\input{X_suppl}
\end{document}

%% file: X_suppl.tex
\clearpage
\appendix
\setcounter{page}{1}
\maketitlesupplementary


\section{Additional Experimental Results}
\label{app:more results on mitigation}
\subsection{Applying Our Method To Stable Diffusion~v1.4}
\begin{figure}[h]
    \centering
    \includegraphics[width=0.45\textwidth]{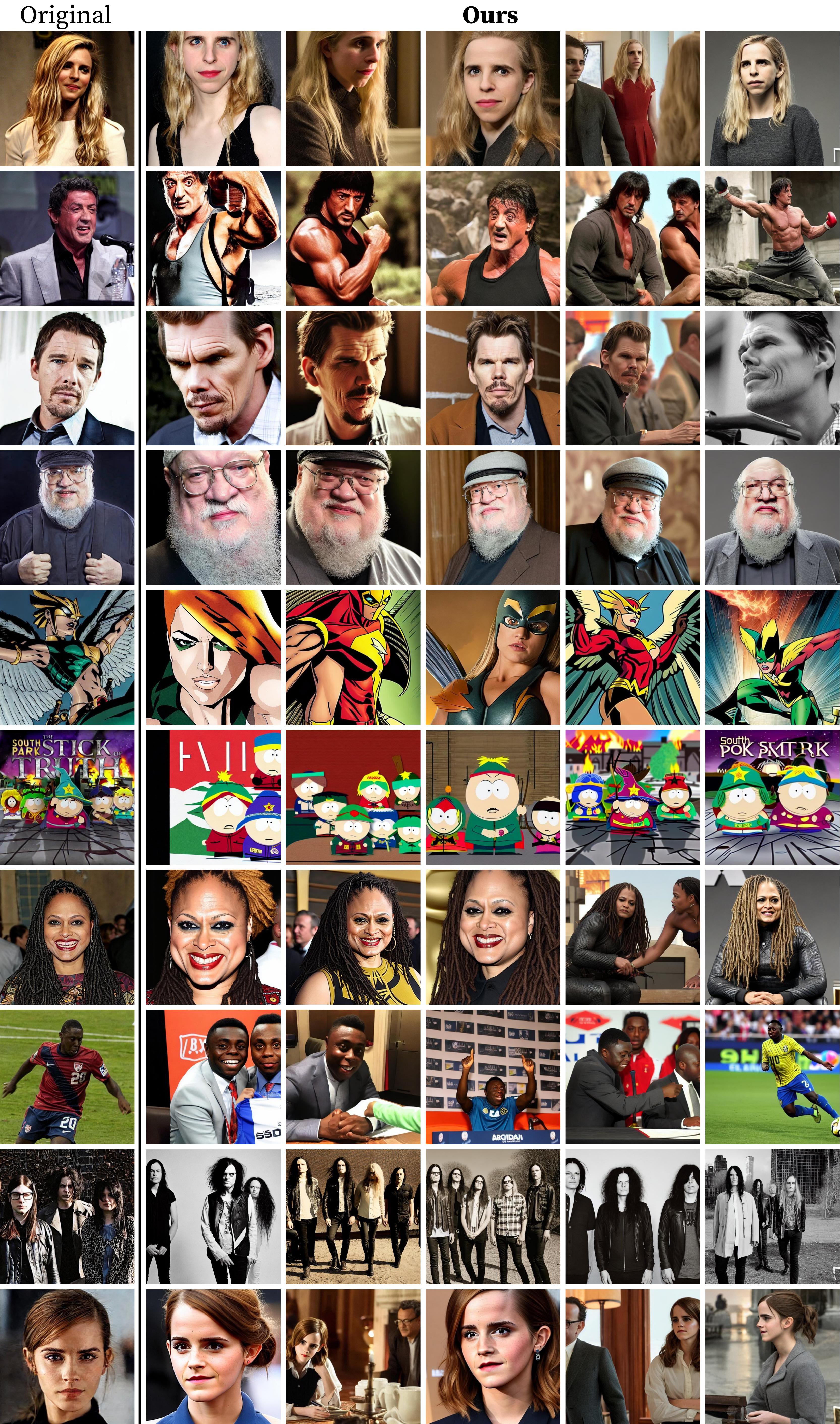}
\caption{\textbf{\pad replacement and $\bE$ masking.} The first column (Original) shows the memorized image consistently reproduced from the original embedding regardless of seeds. 
The remaining five columns (Ours) are generated using our mitigation method with five different random seeds.}
    \label{app:fig:additional res}
\end{figure}

We apply our mitigation method, consisting of \pad replacement and $\bE$ masking. As shown in Figure~\ref{app:fig:additional res}, our method significantly reduces memorization without degrading image quality.
\subsection{Comparison with Prior Mitigation Methods} \label{app:comparison}
\begin{figure}[h]
    \centering
    \includegraphics[width=0.42\textwidth]{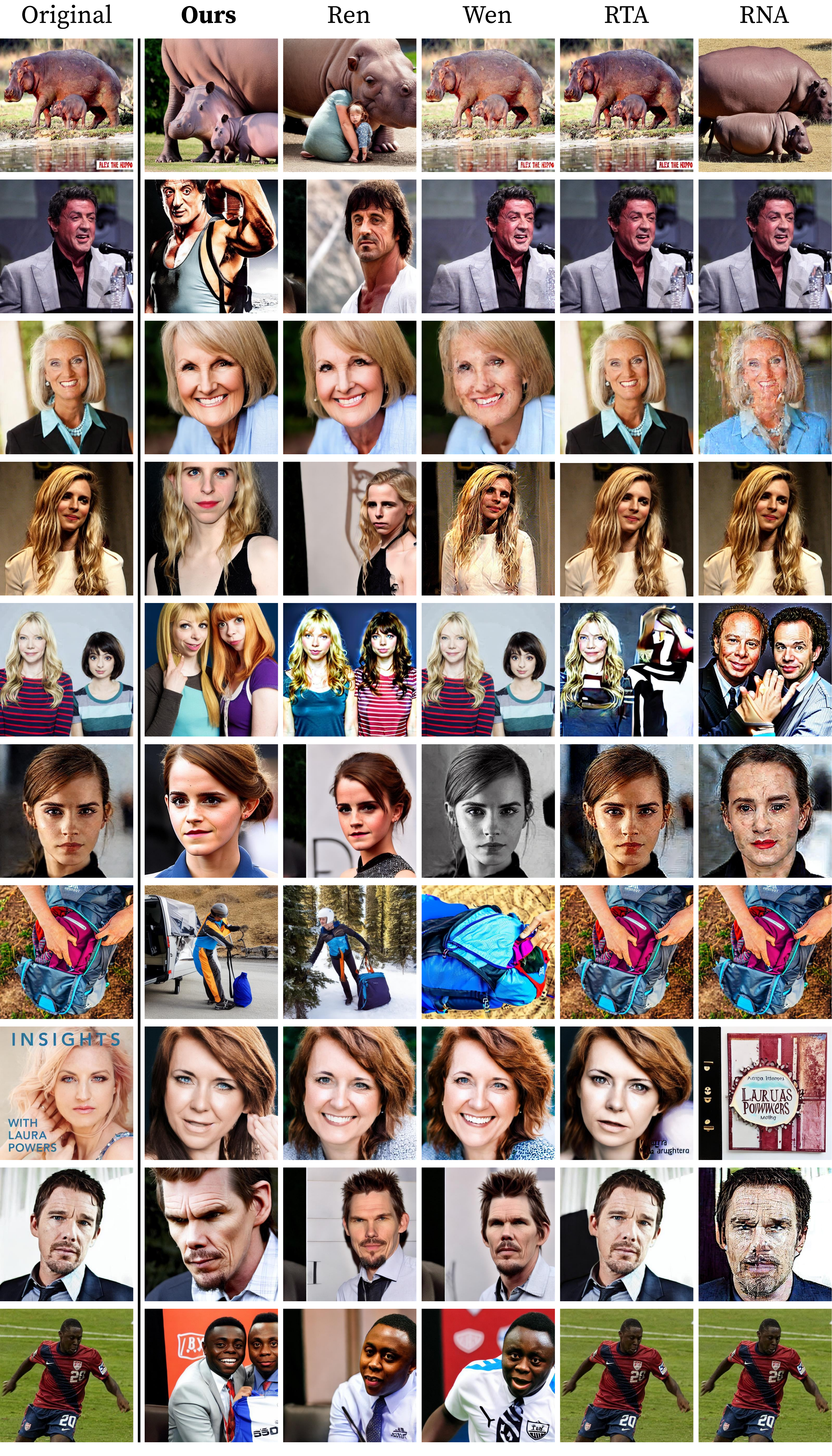}
    \caption{Each row shows images generated from the same prompt using different mitigation methods. Visually and quantitatively, our method produces a more favorable quality-preserving trade-off.}
    \label{app:fig:comparison}
\end{figure}

\paragraph{Comparison with Baselines} 
(1) \citet{ren2024unveiling}, which rescales cross-attention.
(2) \citet{wen2024detecting}, which minimize the magnitude of text-conditional noise prediction. 
(3) Random Token Addition (RTA)~\citep{somepalli2023understanding}, which inserts randomly sampled tokens.
(4) Random Number Addition (RNA)~\citep{somepalli2023understanding}, which inserts a random number between $[0,10^6]$.
We use their optimal configurations \texttt{optim\_target\_loss = 3} for~\citet{wen2024detecting} and \texttt{rescale\_attention = 1.25} for~\citet{ren2024unveiling}.
Results are shown in Figure~\ref{app:fig:comparison}.
\paragraph{Inference Time Comparison.}
As shown in Table~\ref{tab:inference_time}, our mitigation method achieves the fastest inference among all evaluated approaches. 
Since it does not require any additional optimization or computation during inference, its runtime closely matches that of standard text-to-image generation without mitigation. 
In contrast, other methods incur additional computational overhead, resulting in slower inference.
\begin{table}[h]
\centering
\begin{tabular}{lc}
\toprule
\textbf{Method} & \textbf{Inference Time (s)} \\
\midrule
Ours & 3.17 $\pm$ 0.01 \\
\citet{ren2024unveiling} & 3.37 $\pm$ 0.02 \\
\citet{wen2024detecting} & 3.55 $\pm$ 0.16 \\
RNA \citep{somepalli2023understanding} & 3.26 $\pm$ 0.02 \\
RTA \citep{somepalli2023understanding} & 3.31 $\pm$ 0.02 \\
\bottomrule
\end{tabular}
\caption{Average inference time per image for each method.}
\label{tab:inference_time}
\vspace{-2.5em}
\end{table}

\paragraph{Evaluation on Webster 500 Prompts.}
Many prior memorization studies~\citep{wen2024detecting,ren2024unveiling,chenexploring,kowalczuk2025finding} report results directly on all 500 candidate prompts released by \citet{webster2023reproducible}.
In the main paper, we use a stricter curated benchmark of 458 MV prompts, constructed from Webster~\citep{webster2023reproducible} and Membench~\citep{hongmembench}, for primary analysis, as the full 500-prompt set includes unstable and degraded cases that can confound mechanistic interpretation. 
At the same time, to facilitate direct comparison with this widely used evaluation setting, we additionally evaluate all mitigation methods on the full Webster 500 set.

\begin{table}[h]
\centering
\begin{tabular}{lccc}
\toprule
\textbf{Method} & \textbf{SSCD} & \textbf{CLIPScore} & \textbf{Aesthetic} \\
\midrule
Ours & 0.39 $\pm$ 0.20 & 0.29 $\pm$ 0.03 & 5.33 $\pm$ 0.26 \\
Wen & 0.65 $\pm$ 0.22 & 0.31 $\pm$ 0.03 & 5.35 $\pm$ 0.23 \\
Ren & 0.20 $\pm$ 0.15 & 0.26 $\pm$ 0.03 & 4.90 $\pm$ 0.31 \\
RTA & 0.70 $\pm$ 0.23 & 0.31 $\pm$ 0.03 & 5.36 $\pm$ 0.23 \\
RNA & 0.70 $\pm$ 0.23 & 0.32 $\pm$ 0.03 & 5.33 $\pm$ 0.25 \\
\bottomrule
\end{tabular}
\caption{Comparison of mitigation methods on all 500 candidate prompts from Webster~\citep{webster2023reproducible}.}
\label{tab:webster500_comparison}
\vspace{-1em}
\end{table}

As shown in Table~\ref{tab:webster500_comparison}, our method (\pad $\rightarrow$ \texttt{!} \& $\bE \rightarrow \mathbf{0}$) continues to provide a favorable trade-off between memorization reduction and generation quality on this broader benchmark.
In particular, compared to \citet{wen2024detecting}, RNA~\citep{somepalli2023understanding}, and RTA~\citep{somepalli2023understanding}, our method achieves substantially lower SSCD while maintaining comparable CLIPScore and Aesthetic Score.
Although \citet{ren2024unveiling} attains lower SSCD, it does so at a noticeably larger cost in prompt alignment and visual quality.
These results are consistent with our main-paper findings: even on the noisier Webster 500 set, our method remains a competitive mitigation method with a strong quality-preserving trade-off.

\paragraph{Evaluation on Membench.}
To ensure the robustness of our findings, we conduct extensive evaluations across multiple datasets. 
As shown in Table~\ref{tab:comparison membench}, on the Membench~\citep{hongmembench}, which consists of 3,000 prompts, our method shows consistently comparable or better results against existing baselines.
We do not report RNA or RTA results, as their performance is not comparable to these three representative baselines.
\begin{table}[h]
\centering
\begin{tabular}{lccc}
\toprule
\textbf{Method} & \textbf{SSCD} & \textbf{CLIPScore} & \textbf{Aesthetic} \\
\midrule
Ours & 0.16 $\pm$ 0.23 & 0.30 $\pm$ 0.07 & 5.14 $\pm$ 0.53 \\
Ren & 0.15 $\pm$ 0.19 & 0.29 $\pm$ 0.06 & 5.17 $\pm$ 0.51 \\
Wen & 0.79 $\pm$ 0.37 & 0.31 $\pm$ 0.06 & 5.22 $\pm$ 0.48 \\
\bottomrule
\end{tabular}
\caption{Comparison of mitigation methods on Membench~\citep{hongmembench}.}
\label{tab:comparison membench}
\vspace{-1em}
\end{table}

\subsection{Cross-Prompt Swapping Experiment}
\begin{figure}[h]
    \centering
    \begin{minipage}[t]{0.4\linewidth}
        \centering
        \includegraphics[width=\linewidth]{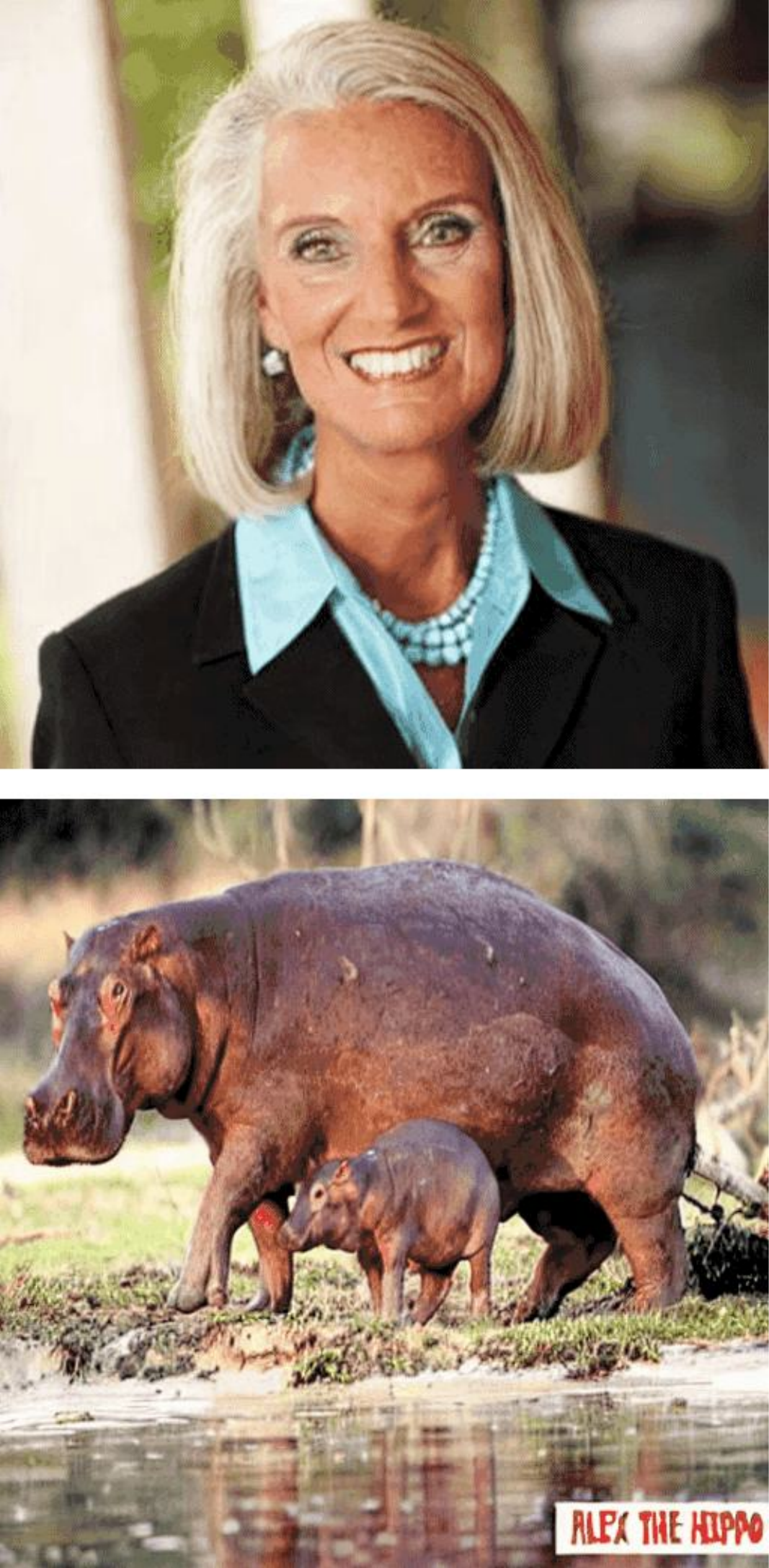}\\
        \textbf{(a)}
    \end{minipage}
    \hspace{1em}
    \begin{minipage}[t]{0.4\linewidth}
        \centering
        \includegraphics[width=\linewidth]{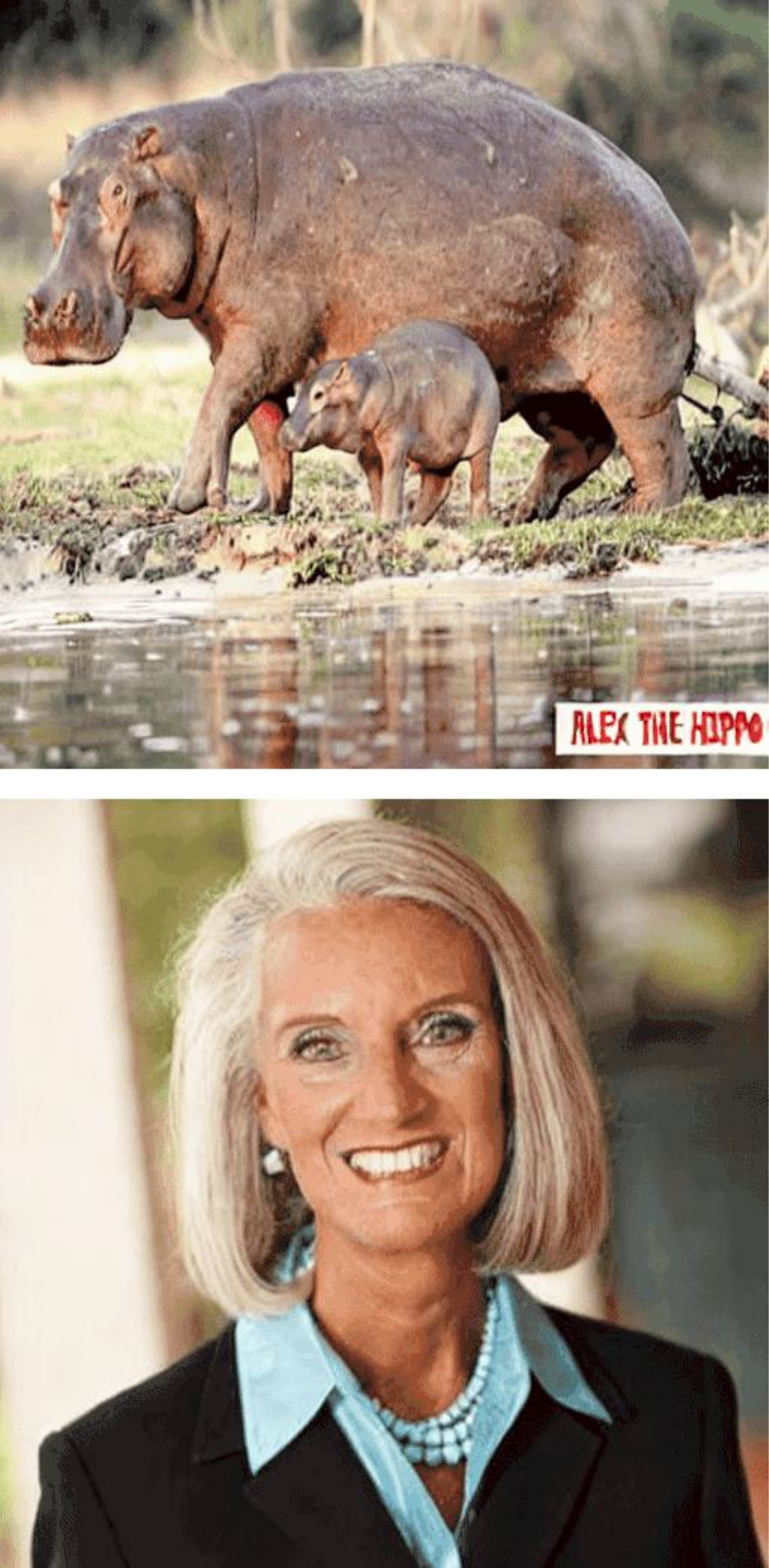}\\
        \textbf{(b)}
    \end{minipage}
    \caption{(a) Original generation without swapping. 
    (b) Generation after swapping both $\bE$ and $\bp_i$, showing that the output changes to reflect the swapped prompt.}
    \label{app:swap}
\end{figure}

To further investigate the role of $\bE$, we conduct an experiment swapping $\bE$ and $\bp_i$
between different prompts.
Our experiment reveals that swapping only $\bE$ has no discernible effect on the output,
whereas swapping both $\bE$ and $\bp_i$ alters the generated image to match 
the swapped prompt (Figure~\ref{app:swap}).
This suggests that $\bp_i$, much like $\bE$, encode critical semantic information rather than serving as mere placeholders.
\subsection{Robustness of our mitigation method on non-memorized prompts} \label{app:non}

\begin{figure*}[t]
    \centering
    \begin{minipage}[t]{0.23\textwidth}
        \centering
        \includegraphics[width=\textwidth]{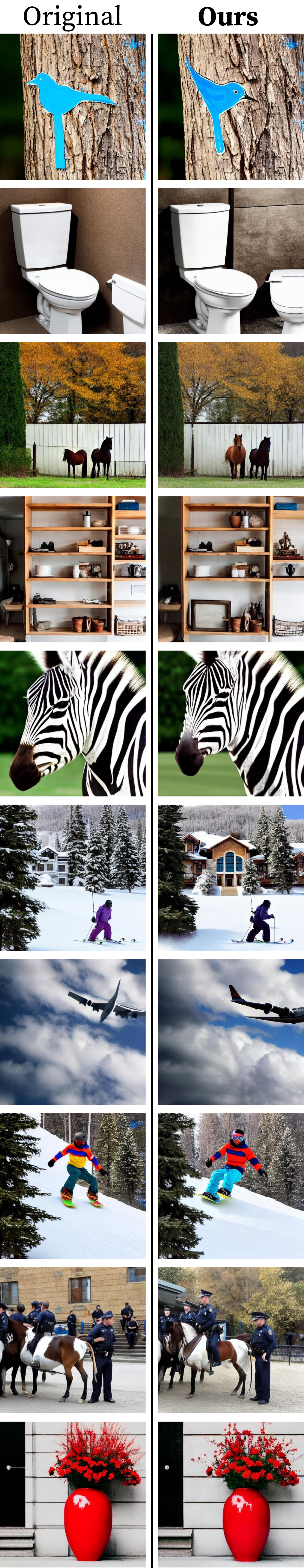}
        \vspace{0.5em}
        \textbf{(a)} MS COCO
    \end{minipage}
    \hspace{1em}
    \begin{minipage}[t]{0.23\textwidth}
        \centering
        \includegraphics[width=\textwidth]{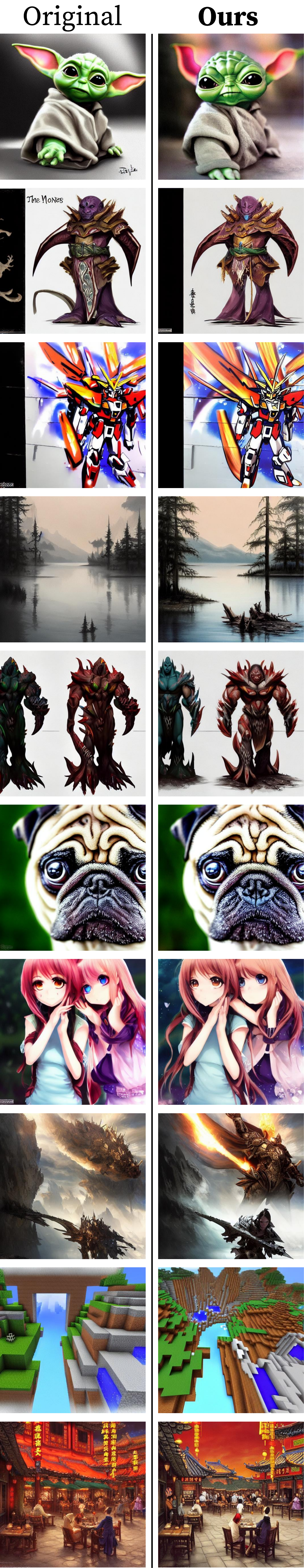}
        \vspace{0.5em}
        \textbf{(b)} Lexica Art
    \end{minipage}
    \hspace{1em}
    \begin{minipage}[t]{0.23\textwidth}
        \centering
        \includegraphics[width=\textwidth]{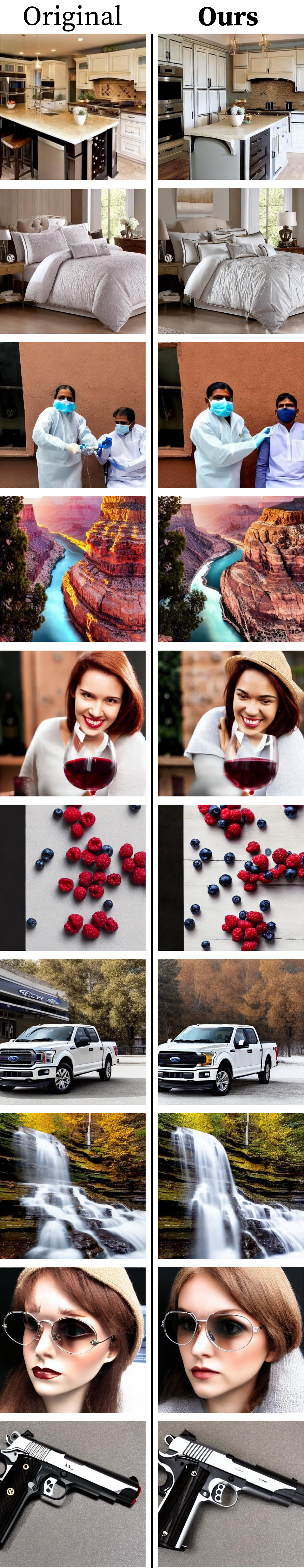}
        \vspace{0.5em}
        \textbf{(c)} LAION
    \end{minipage}
    \caption{ ``Original'' is generated using the default embeddings, while ``Ours'' applies our mitigation strategy (\pad $\rightarrow$ \texttt{!} \& $\bE \rightarrow \mathbf{0}$).}
    \label{app:fig:nonmem}
\end{figure*}

We utilize non-memorized prompts to evaluate whether our mitigation method impacts general generation quality.
Following the prior studies~\citep{wen2024detecting, chenexploring}, we sample 500 prompts each from MS COCO, Lexica Art, and LAION. To ensure a fair and rigorous comparison, we apply additional filtering so that the selected prompts resemble memorized prompts in structure and complexity: each prompt must consist only of ASCII characters; exclude URLs (e.g., “http”, “www”); and have a tokenized length between 5 and 40.

As shown in Figure~\ref{app:fig:nonmem}, our mitigation method preserves visual quality and prompt alignment across a diverse range of non-memorized prompts, demonstrating robustness and deployability without degradation in general generation. 


\section{Detailed Experimental Settings}
\label{app:setup}
\paragraph{Model}
We perform all experiments using Stable Diffusion, specifically v1.4 for the SD~1 series and v2.1 for the SD~2 series.  
Image generation is carried out with the DDIM scheduler~\citep{songdenoising}, using a guidance scale of 7.5 and 50 inference steps, following prior memorization works~\citep{somepalli2023understanding, wen2024detecting}.
Detailed software libraries and configurations are provided in the Supplementary Material.

\paragraph{Computational Resources}
All experiments were conducted on a single NVIDIA L40S GPU (48 GB VRAM) and an AMD EPYC 9754 processor (192 CPU cores).
Because our study focuses exclusively on inference rather than training, the overall computational requirements were modest, and the full experimental pipeline ran efficiently on a single GPU.
\paragraph{Dataset and Evaluation}
To build the dataset used in our experiments, we apply filtering criteria to a subset of Membench~\citep{hongmembench} prompts, using images generated with 10 fixed random seeds: {0, 1, 10, 42, 100, 441, 515, 1000, 2025, 10000} (see Section~\ref{sec:re-evaluating}), and average the results over them to ensure stability and comparability.

\subsection{Analysis of Memorized Prompt Datasets from Prior Works}
\label{app:dataset analysis}
\begin{figure}[h]
    \centering
    \begin{minipage}[t]{0.2\textwidth}
        \centering
        \includegraphics[width=\textwidth]{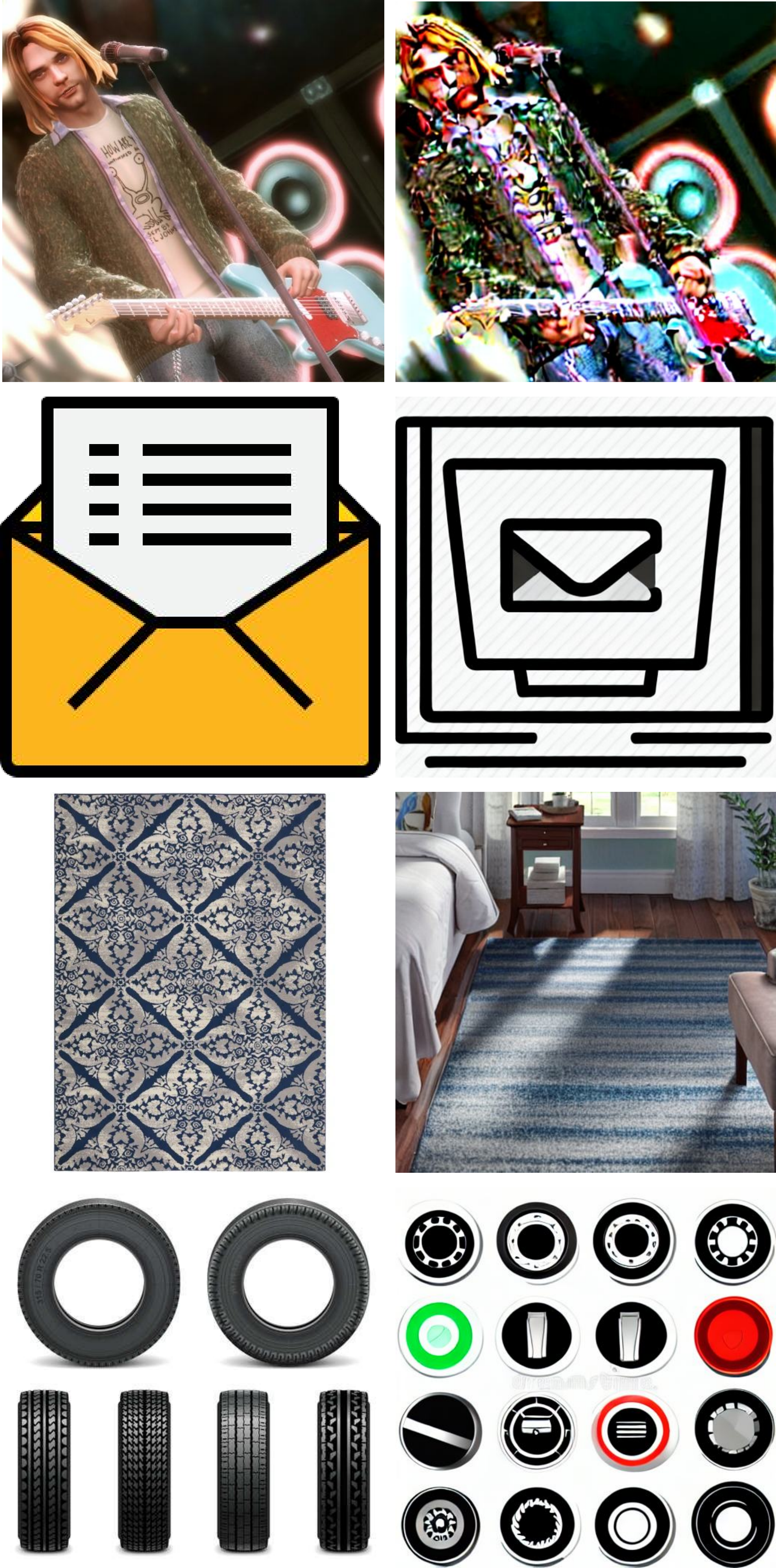}
    \end{minipage}
    \hspace{1em}
    \begin{minipage}[t]{0.2\textwidth}
        \centering
        \includegraphics[width=\textwidth]{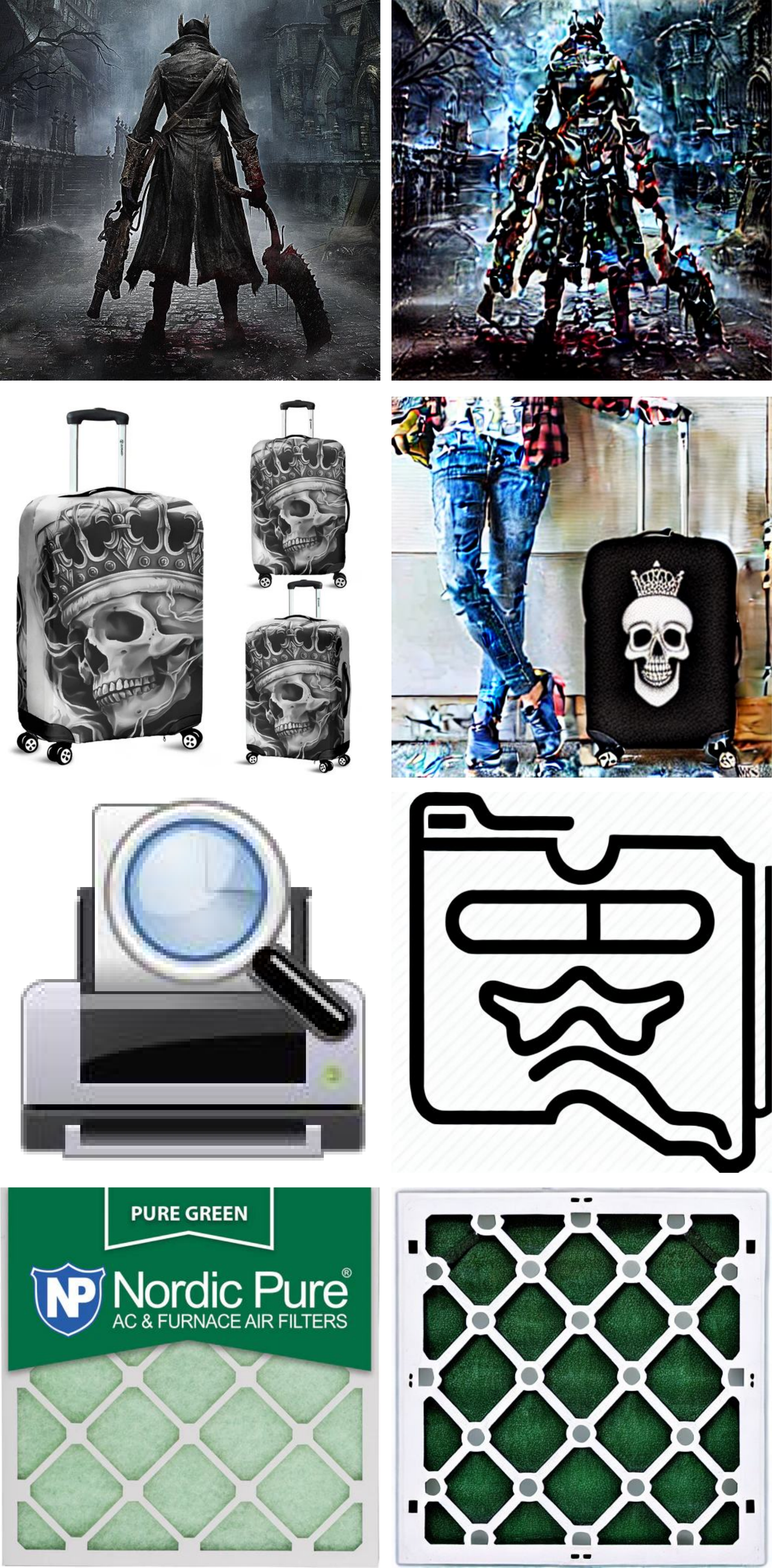}
    \end{minipage}
    \caption{Each left column shows the reference image, and each right column shows the corresponding model-generated image.}
    \label{app:fig:not_valid}
\end{figure}

\citet{webster2023reproducible} released 500 candidate memorized prompts for Stable Diffusion~v1.4 (SD1).
However, as we discussed earlier, a considerable portion of these prompts are not truly \emph{memorized} cases.
As shown in Figure~\ref{app:fig:not_valid}, many of them produce degraded or entirely unrelated images,
and in some cases, the generated image corresponds to a completely different prompt.
Therefore, directly using all 500 prompts without verification would be inappropriate.

\citet{webster2023reproducible} further categorized 345 prompts for SD1 into 86 \emph{Matching Verbatim} (MV), 30 \emph{Retrieval Verbatim} (RV), and 219 \emph{Template Verbatim} (TV) cases. 
For Stable Diffusion~v2.1 (SD2), the same study includes 219 prompts, with only 4 labeled as MV and 215 as TV.

\citet{hongmembench} provided 3000 memorized prompts for SD1 without further categorization. Prompts corresponding to SD2 are not publicly available.

To better understand the characteristics of memorized prompts, we analyze the token length statistics of prompts used in prior memorization studies. Table~\ref{tab:prompt_token_statistics} summarizes the statistics for SD1 and SD2 prompts reported in prior memorization studies.

\begin{table*}[h]
\centering
\begin{tabular}{lccc}
\toprule
\textbf{Metric} & \multicolumn{2}{c}{\textbf{SD1}} & \textbf{SD2} \\
\cmidrule(lr){2-3}
 & \citet{webster2023reproducible} & \citet{hongmembench} & \citet{webster2023reproducible} \\
\midrule
Number of Prompts & 345 & 3000 & 219 \\
Mean Length  & 19.74 & 15.31 & 21.51 \\
Max / Min Length  & 38 / 4 & 38 / 5 & 35 / 6 \\
\bottomrule
\end{tabular}
\caption{Prompt Token Length Statistics for Stable Diffusion v1.4 (SD1) and Stable Diffusion v2.1 (SD2)}
\label{tab:prompt_token_statistics}
\end{table*}

\subsection{Prompts used for Figures}
To enhance reproducibility and transparency, we list the exact prompts used for generating the example images shown in our figures. 
The memorized prompts are drawn from publicly released datasets provided by~\citet{webster2023reproducible} and~\citet{hongmembench}, 
while the non-memorized prompts are sampled from MS COCO~\citep{lin2014microsoft}, Lexica~\citep{santana_gustavostastable-diffusion-prompts_2022} and LAION~\citep{schuhmann2022laion}.
\paragraph{Figure~\ref{fig:main}} 
\begin{itemize}
\item Living in the Light with Ann Graham Lotz
\item Mothers influence on her young hippo
\item Here's Who Ian McShane May Be Playing in \textless i\textgreater Game of Thrones\textless /i\textgreater{} Season Six
\end{itemize}

\paragraph{Figure~\ref{fig:prompt intervention}} 
\begin{itemize}
\item Living in the Light with Ann Graham Lotz
\item Talks on the Precepts and Buddhist Ethics
\end{itemize}

\paragraph{Figure~\ref{fig:prompt embedding}} 
\begin{itemize}
\item Mothers influence on her young hippo
\item The No Limits Business Woman Podcast
\end{itemize}

\paragraph{Figure~\ref{fig:comparison}} 
\begin{itemize}
\item Björk Explains Decision To Pull \textless i\textgreater Vulnicura\textless /i\textgreater{} From Spotify
\item "As Punisher Joins \textless i\textgreater Daredevil\textless /i\textgreater{} Season Two, Who Will the New Villain Be?"
\item Talks on the Precepts and Buddhist Ethics
\end{itemize}

\paragraph{Figure~\ref{fig:partially masking}} 
\begin{itemize}
\item Emma Watson to play Belle in Disney's \textless i\textgreater Beauty and the Beast\textless /i\textgreater
\item "\textless i\textgreater The Long Dark\textless /i\textgreater{} Gets First Trailer, Steam Early Access"
\item "Long-Lost F. Scott Fitzgerald Story Rediscovered and Published, 76 Years Later"
\end{itemize}

\paragraph{Figure~\ref{app:fig:additional res}} 
\begin{itemize}
\item Brit Marling-Zal Batmanglij Drama Series \textless i\textgreater The OA\textless /i\textgreater{} Gets Picked Up By Netflix
\item Sarah Silverman Will Star in HBO Pilot from \textless i\textgreater Secret Diary of a Call Girl\textless /i\textgreater{} Creator
\item Ethan Hawke to Star as Jazz Great Chet Baker in New Biopic
\item "George R.R. Martin to Focus on Writing Next Book, World Rejoices"
\item Hawkgirl Cast in \textless i\textgreater Arrow\textless /i\textgreater{}/\textless i\textgreater Flash\textless /i\textgreater{} Spinoff Series For The CW
\item \textless em\textgreater South Park: The Stick of Truth\textless /em\textgreater{} Review (Multi-Platform)
\item Ava DuVernay Won't Direct \textless i\textgreater Black Panther\textless /i\textgreater{} After All
\item Freddy Adu Signs For Yet Another Club You Probably Don't Know
\item """""""Listen to The Dead Weather's New Song, """"""""Buzzkill(er)"""""""""""""""
\item Emma Watson Set to Star Alongside Tom Hanks in Film Adaptation of Dave Eggers' \textless i\textgreater The Circle\textless /i\textgreater
\end{itemize}

\paragraph{Figure~\ref{app:fig:comparison}}
\begin{itemize}
\item Mothers influence on her young hippo
\item Rambo 5 und Rocky Spin-Off - Sylvester Stallone gibt Updates
\item Living in the Light with Ann Graham Lotz
\item Brit Marling-Zal Batmanglij Drama Series \textless i\textgreater The OA\textless /i\textgreater{} Gets Picked Up By Netflix
\item Watch the First Episode of \textless i\textgreater Garfunkel and Oates\textless /i\textgreater{}
\item Emma Watson Set to Star Alongside Tom Hanks in Film Adaptation of Dave Eggers' \textless i\textgreater The Circle \textless /i\textgreater{}
\item Full body U-Zip main opening - Full body U-Zip main opening on front of bag for easy unloading when you get to camp
\item Insights with Laura Powers
\item Ethan Hawke to Star as Jazz Great Chet Baker in New Biopic
\item Freddy Adu Signs For Yet Another Club You Probably Don't Know
\end{itemize}

\paragraph{Figure~\ref{app:fig:nonmem}}

\begin{itemize}
\item [] \textit{MS COCO}
\begin{itemize}
\item A painted sign of a blue bird in a tree in the woods.
\item a small white toilet is in a tiny bathroom
\item Two horses standing behind fence with grassy leaves
\item A lot of wooden shelves filled with lots of clutter.
\item A zebra standing on the grass holding its head near the ground.
\item a skier and a snowboarder in front of a large house
\item Large airplane flying below the clouds from underneath
\item A snowboarder in a colorful jacket racing down a slope.
\item There are police men standing and some are sitting on horses
\item A large red vase sitting in front of a building. 
\end{itemize}
\end{itemize}

\begin{itemize}
\item [] \textit{Lexica Art}
\begin{itemize}
\item Baby Yoda by Jeszika Le Vye, trending on artstation, hyperdetalied,
\item the dnd monster pung\_ concept art
\item graffiti on a wall of anime gundam, speed painting, trending on artstation
\item A beautiful concept art painting of a gloomy lake with a body floating slightly below the surface by
\item three different views of a predator set for dota 2, concept art by senior character artist, trending on artstation, artstation hd, full body
\item closeup of a pug with moss growing out of its face folds, macro photography, overgrown pug, high resolution photo, trending on artstation
\item two beautiful idols standing face to face, trending artstation, pixiv, detailed anime art
\item guild wars 2, cinematic battlefield, Hyperrealistic CGI Photorealistic cinematic volume lighting fanart on ArtStation full sun shine day concept art, digital art, high detail by Daniel Dociu
\item minecraft style concept art of a blue portal to a fantasy world
\item fantasy art of a bustling tavern in china, at night, by justin gerard, highly detailed digital art, trending on artstation
\end{itemize}
\end{itemize}

\begin{itemize}
\item [] \textit{LAION}
\begin{itemize}
\item kitchen ideas with islands simply home designs home design ideas 3
\item 78 most terrific supple duvet sets with matching let curtains covers curtain duvets cover definition define snazzy aequorea plus thresh expensive quilts
\item India begins next phase of COVID vaccination from Monday:
\item Gary Hart Photography: Bright Angel Lightning, Grand Canyon
\item Beautiful happy lady drinking glass of red wine.
\item berries\_lead
\item 2018 Ford F-150
\item Alpine Falls in Loyalsock State Forest
\item Retro Classic Full Metal Tear Drop Clear Lens Aviator Glasses C936
\item Well M1911 A1 Vollmetall Springer 6mm BB Two-Tone
\end{itemize}
\end{itemize}

\paragraph{Figure~\ref{app:fig:not_valid}}
\begin{itemize}
\item 33 Screenshots of Musicians in Videogames
\item ``communication, email, mail, message, online, open icon"
\item Anzell Blue/Gray Area Rug by Andover Mills
\item Vector Tyre Icons isolated on white background Illustration
\item \textless em\textgreater{}Bloodborne\textless/em\textgreater{} Video: Sony Explains the Game's Procedurally Generated Dungeons
\item 3D Black \& White Skull King Design Luggage Covers 007
\item ``document,print,preview icon"
\item 12x12x1 Pure Green AC Furnace Air Filters Qty 6 - Nordic Pure
\end{itemize}

\clearpage

\section{Difference of embeddings between CLIP and OpenCLIP}\label{app:clip and openclip}

In the CLIPText encoder used in Stable Diffusion~v1.4, text embeddings are grouped into three clusters: $\bS$, $\bt_i$, and a merged cluster of $\bE$ and $\bp_i$. 
This is because the tokenizer duplicates the \eot for padding, making $\bp_i$ nearly identical to $\bE$ after encoding.
In contrast, the CLIPText encoder of Stable Diffusion~v2.1 yields a separate cluster for $\bE$ and $\bp_i$, indicating a semantic distinction between the two.

Stable Diffusion~v1.4 employs the original CLIP text encoder, where the tokenizer uses \eot as the default \pad. As a result, $\bp_i$ are near duplicates of $\bE$, which amplifies the semantic influence of $\bE$ and contributes significantly to memorization.

Stable Diffusion~v2.1 employs an OpenCLIP text encoder.
Importantly, OpenCLIP maintains the same overall architecture and embedding mechanism as the original CLIP, 
with only minor differences in training scale and data.
The critical difference relevant to our analysis is that the OpenCLIP tokenizer uses a neutral \texttt{!} token for \pad instead of duplicating \eot.
As a result, $\bp_i$ are semantically distinct from the $\bE$, 
eliminating the structural duplication that contributed to memorization in Stable Diffusion~v1.4.

\begin{figure}[!h]
    \centering
    \begin{minipage}{0.8\linewidth}
        \centering
        \includegraphics[width=\linewidth]{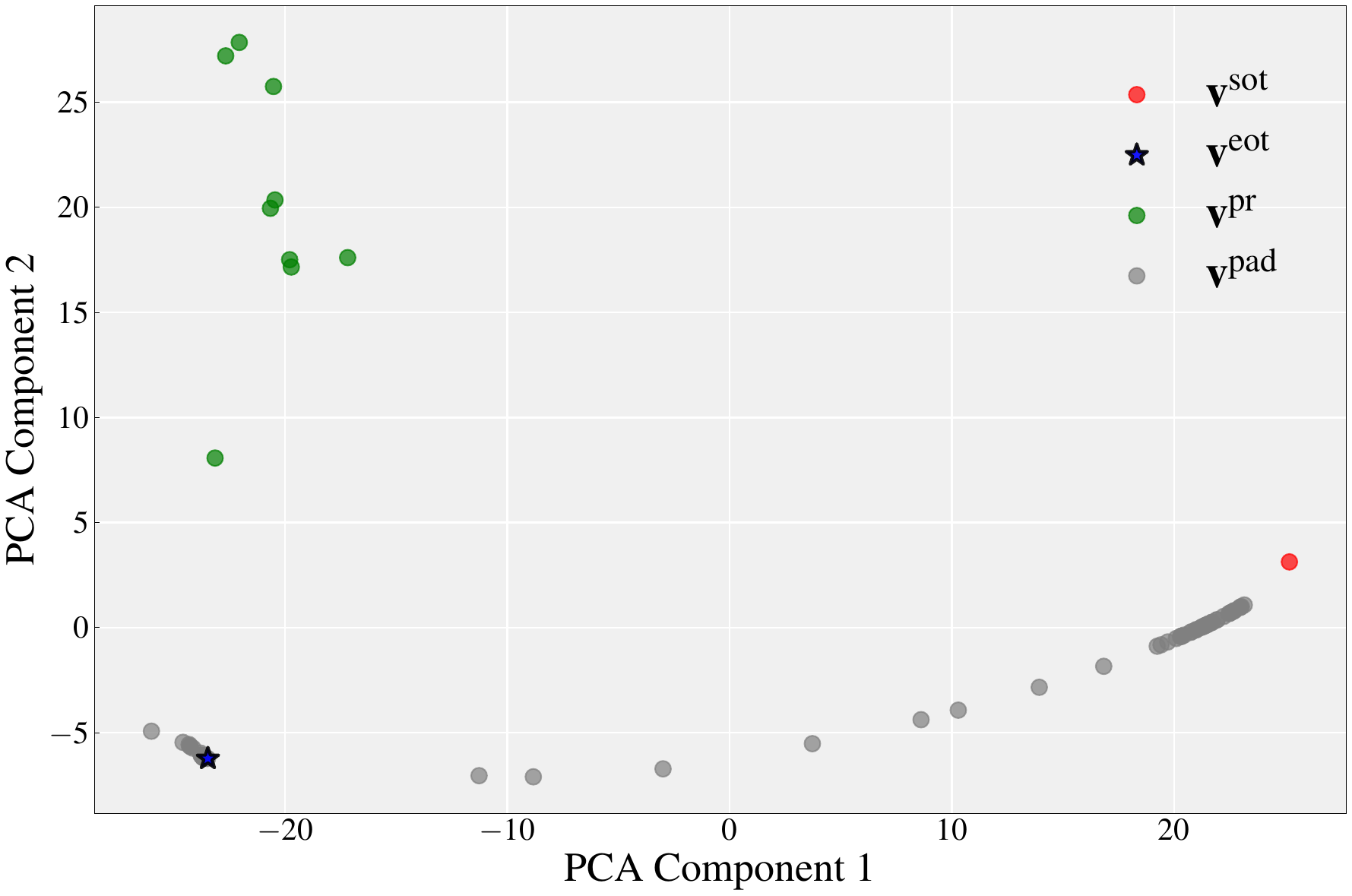}\\
        \includegraphics[width=\linewidth]{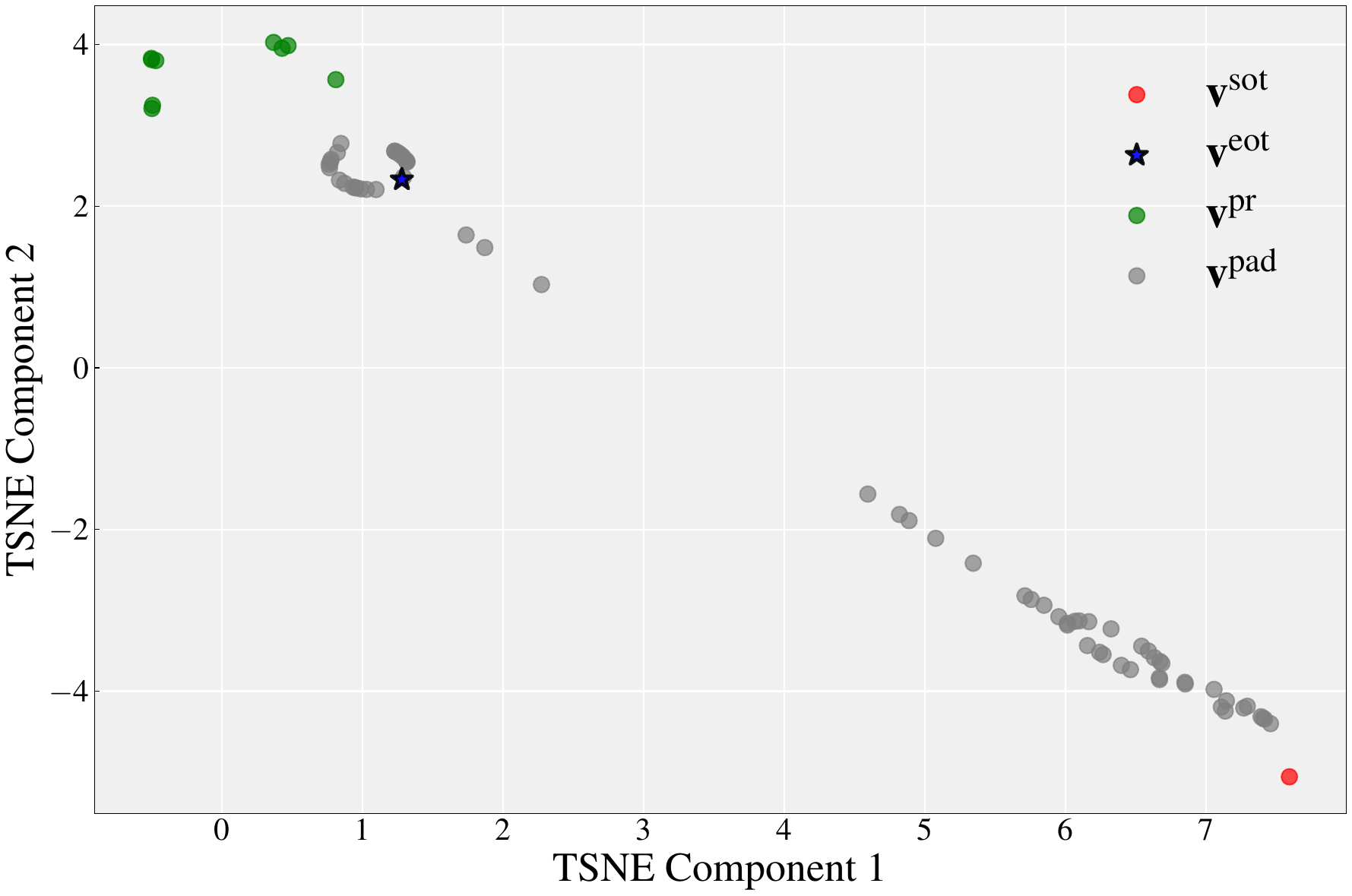}\\
        \textbf{(a)} CLIP: PCA (top) and t-SNE (bottom)
    \end{minipage}\\
    \vspace{1em}
    \begin{minipage}{0.8\linewidth}
        \centering
        \includegraphics[width=\linewidth]{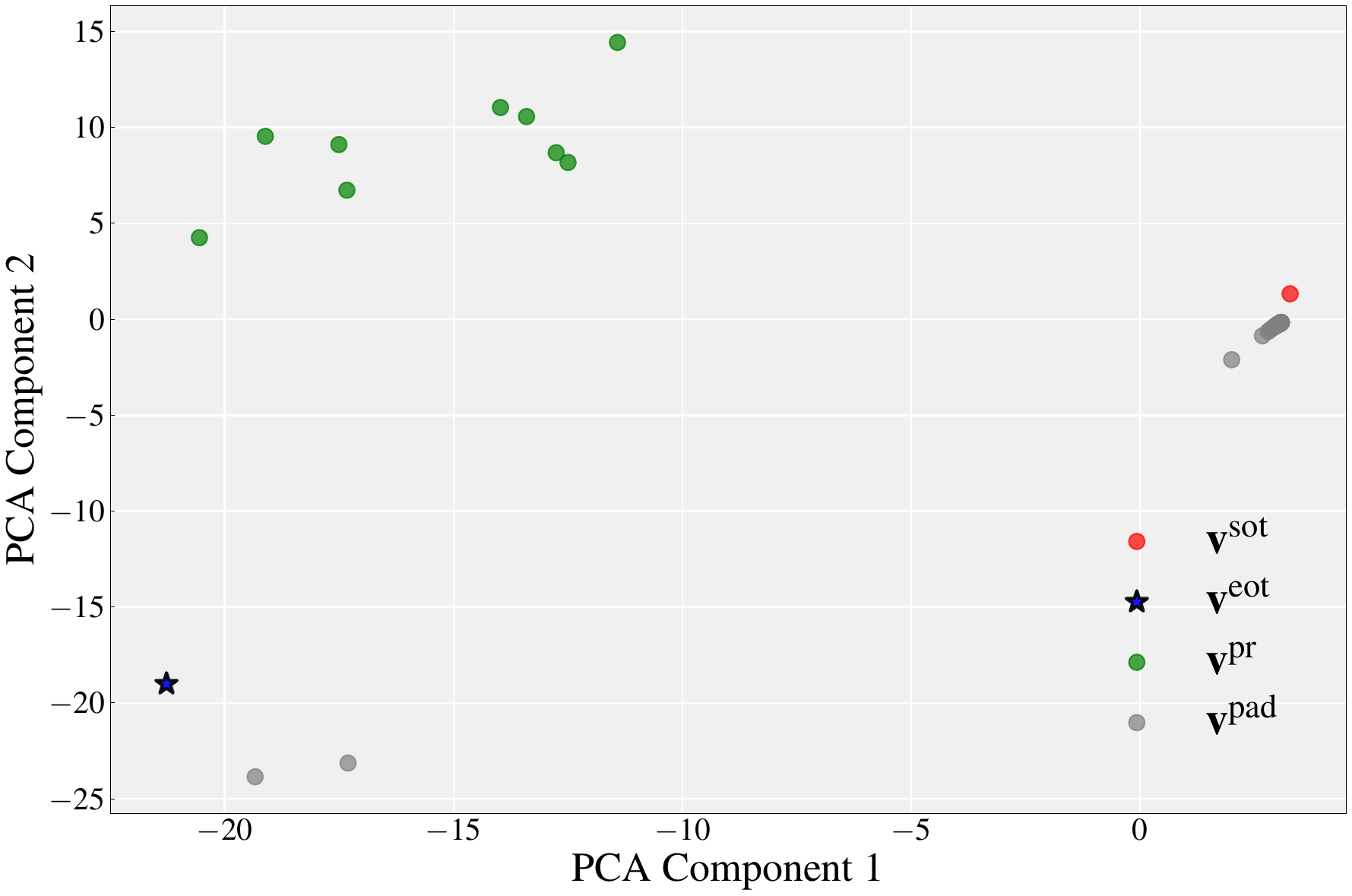}\\
        \includegraphics[width=\linewidth]{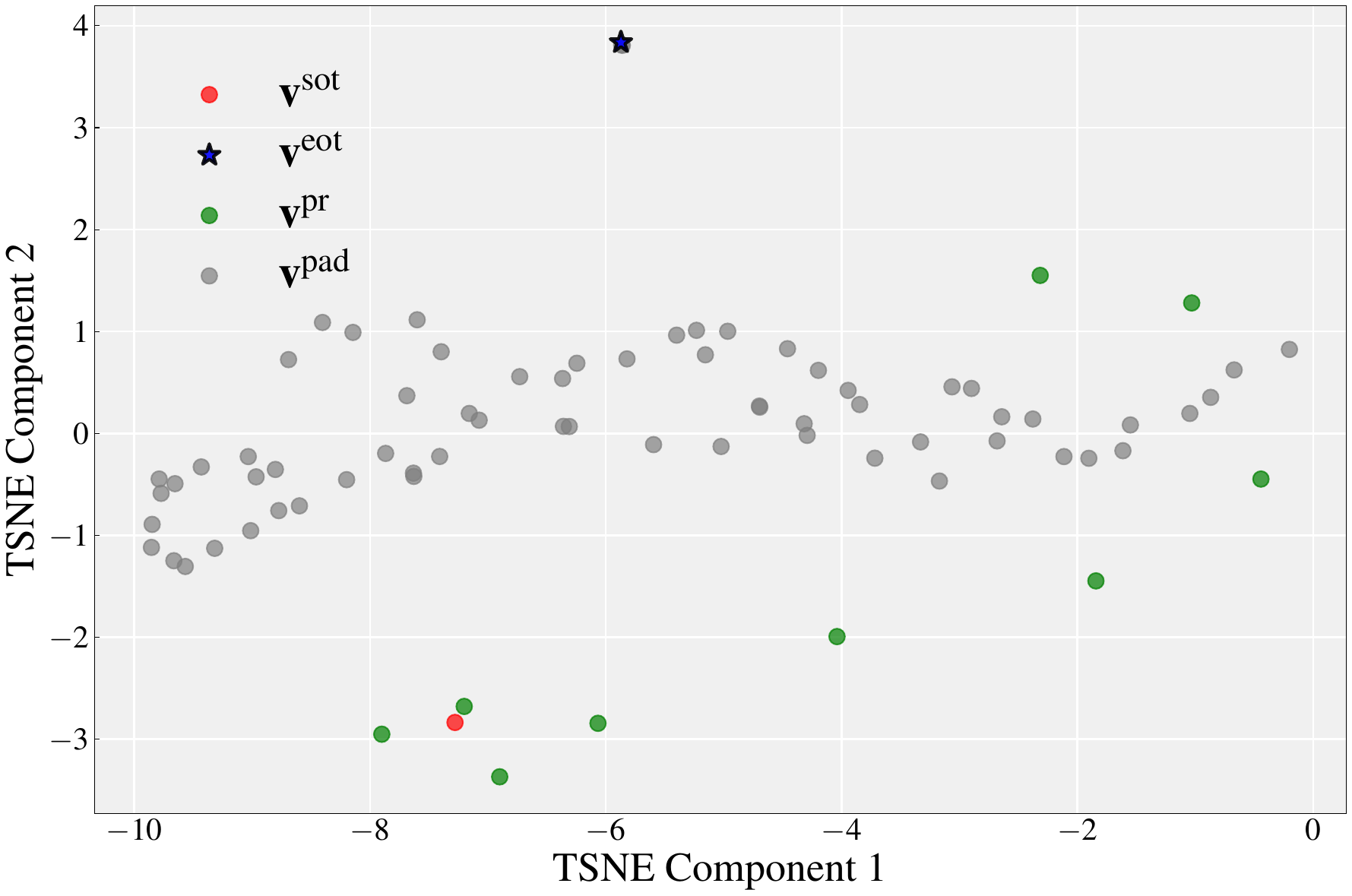}\\
        \textbf{(b)} OpenCLIP: PCA (top) and t-SNE (bottom)
    \end{minipage}
    \caption{Text embeddings of ``Living in the Light with Ann Graham Lotz'' visualized using PCA and t-SNE for (a) CLIP and (b) OpenCLIP models. The two projection methods reveal distinct clustering behaviors.}
    \label{fig:clip_embeddings}
\end{figure}

The PCA and t-SNE visualizations in Figure~\ref{fig:clip_embeddings} further illustrate this difference.



\section{Cross-Attention Maps by Embedding}
\label{app:attn}
\begin{figure*}[h]
    \centering
    \includegraphics[width=0.9\textwidth]{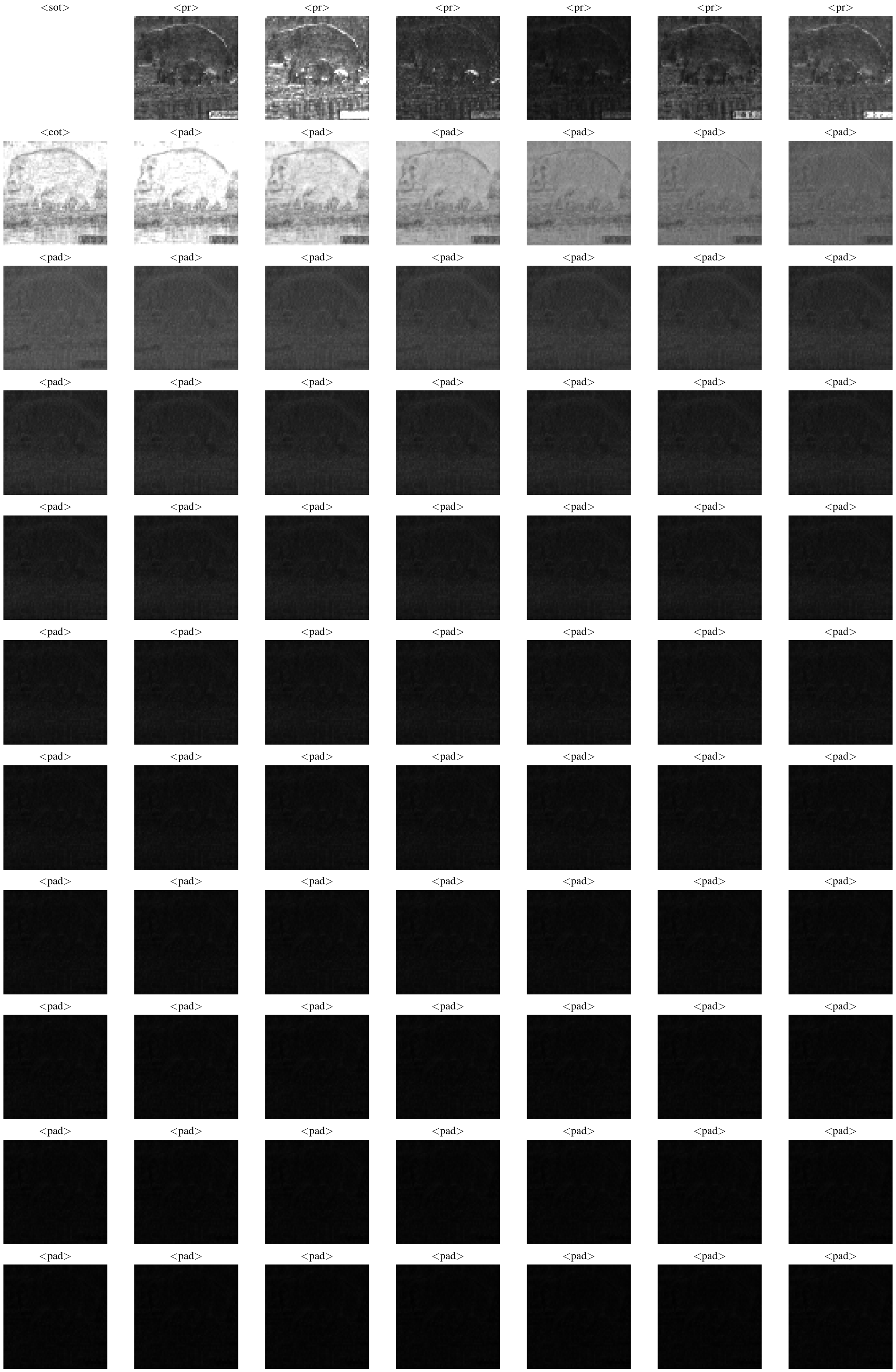}
    \caption{Cross Attention maps of ``Mothers influence on her young hippo''}
    \label{fig:cross_attn_map}
\end{figure*}

\begin{figure*}[h]
    \centering
    \includegraphics[width=0.9\textwidth]{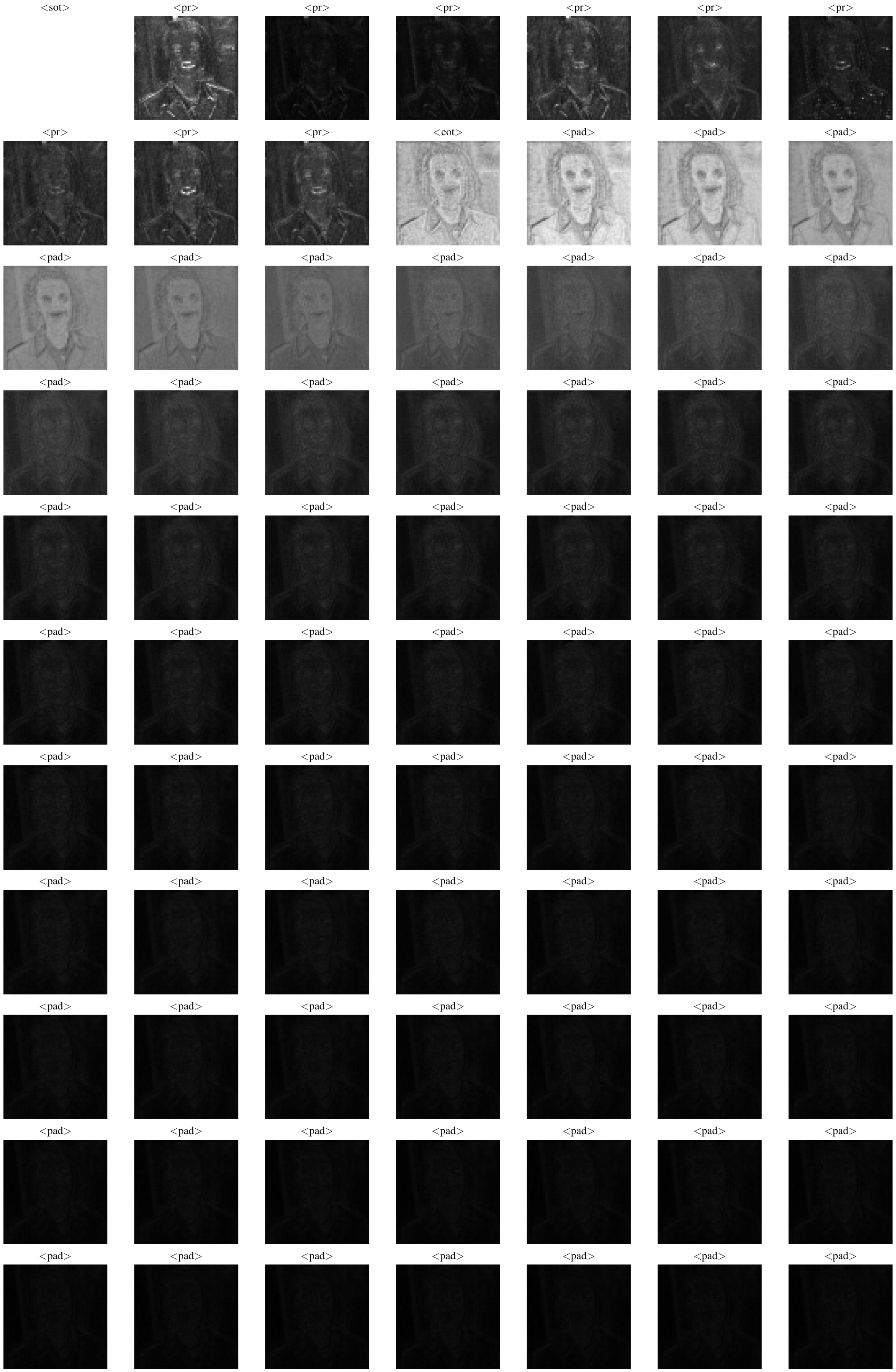}
    \caption{Cross Attention maps of ``Living in the Light with Ann Graham''}
    \label{fig:cross_attn_map_2}
\end{figure*}

Following \citet{chenexploring}, we visualize the cross-attention maps from the first two 64-pixel downsampling layers of the U-Net at the last denoising step. 
To ensure consistent brightness across all maps, we normalize the intensity by setting the minimum and maximum brightness levels equal across all cross-attention maps.
As shown in Figure~\ref{fig:cross_attn_map} and \ref{fig:cross_attn_map_2}, not only does the attention map corresponding to $\bE$ appear bright, but the adjacent $\bp_i$ also exhibit high attention scores.

\clearpage

%% file: main.bib
@String(CVPR= {IEEE Conf. Comput. Vis. Pattern Recog.})

@String(ECCV= {Eur. Conf. Comput. Vis.})

@String(ICLR = {Int. Conf. Learn. Represent.})

@String(CVPR  = {CVPR})

@String(ECCV  = {ECCV})

@String(ICLR  = {ICLR})

@inproceedings{carlini2023extracting,
  title={Extracting training data from diffusion models},
  author={Carlini, Nicolas and Hayes, Jamie and Nasr, Milad and Jagielski, Matthew and Sehwag, Vikash and Tramer, Florian and Balle, Borja and Ippolito, Daphne and Wallace, Eric},
  booktitle={USENIX Security},
  year={2023}
}

@inproceedings{radford2021learning,
  title={Learning transferable visual models from natural language supervision},
  author={Radford, Alec and Kim, Jong Wook and Hallacy, Chris and Ramesh, Aditya and Goh, Gabriel and Agarwal, Sandhini and Sastry, Girish and Askell, Amanda and Mishkin, Pamela and Clark, Jack and others},
  booktitle={ICML},
  year={2021},
}

@inproceedings{rombach2022high,
  title={High-resolution image synthesis with latent diffusion models},
  author={Rombach, Robin and Blattmann, Andreas and Lorenz, Dominik and Esser, Patrick and Ommer, Bj{\"o}rn},
  booktitle={CVPR},
  year={2022}
}

@inproceedings{chenexploring,
  title={Exploring Local Memorization in Diffusion Models via Bright Ending Attention},
  author={Chen, Chen and Liu, Daochang and Shah, Mubarak and Xu, Chang},
  booktitle={ICLR},
  year={2025}
}

@inproceedings{wen2024detecting,
  title={Detecting, explaining, and mitigating memorization in diffusion models},
  author={Wen, Yuxin and Liu, Yuchen and Chen, Chen and Lyu, Lingjuan},
  booktitle={ICLR},
  year={2024}
}

@inproceedings{ren2024unveiling,
  title={Unveiling and mitigating memorization in text-to-image diffusion models through cross attention},
  author={Ren, Jie and Li, Yaxin and Zeng, Shenglai and Xu, Han and Lyu, Lingjuan and Xing, Yue and Tang, Jiliang},
  booktitle={ECCV},
  year={2024},
}

@inproceedings{somepalli2023understanding,
  title={Understanding and mitigating copying in diffusion models},
  author={Somepalli, Gowthami and Singla, Vasu and Goldblum, Micah and Geiping, Jonas and Goldstein, Tom},
  booktitle={NeurIPS},
  year={2023}
}

@inproceedings{pizzi2022self,
  title={A self-supervised descriptor for image copy detection},
  author={Pizzi, Ed and Roy, Sreya Dutta and Ravindra, Sugosh Nagavara and Goyal, Priya and Douze, Matthijs},
  booktitle={CVPR},
  year={2022}
}

@article{webster2023reproducible,
  title={A reproducible extraction of training images from diffusion models},
  author={Webster, Ryan},
  journal={arXiv preprint arXiv:2305.08694},
  year={2023}
}

@inproceedings{schuhmann2022laion,
  title={Laion-5b: An open large-scale dataset for training next generation image-text models},
  author={Schuhmann, Christoph and Beaumont, Romain and Vencu, Richard and Gordon, Cade and Wightman, Ross and Cherti, Mehdi and Coombes, Theo and Katta, Aarush and Mullis, Clayton and Wortsman, Mitchell and others},
  booktitle={NeurIPS},
  year={2022}
}

@inproceedings{hessel2021clipscore,
  title={CLIPScore: A Reference-free Evaluation Metric for Image Captioning},
  author={Hessel, Jack and Holtzman, Ari and Forbes, Maxwell and Le Bras, Ronan and Choi, Yejin},
  booktitle={EMNLP},
  year={2021}
}

@inproceedings{toker2025padding,
  title={Padding tone: A mechanistic analysis of padding tokens in t2i models},
  author={Toker, Michael and Galil, Ido and Orgad, Hadas and Gal, Rinon and Tewel, Yoad and Chechik, Gal and Belinkov, Yonatan},
  booktitle={NAACL},
  year={2025}
}

@inproceedings{cherti2023reproducible,
  title={Reproducible scaling laws for contrastive language-image learning},
  author={Cherti, Mehdi and Beaumont, Romain and Wightman, Ross and Wortsman, Mitchell and Ilharco, Gabriel and Gordon, Cade and Schuhmann, Christoph and Schmidt, Ludwig and Jitsev, Jenia},
  booktitle={CVPR},
  year={2023}
}

@inproceedings{podellsdxl,
  title={SDXL: Improving Latent Diffusion Models for High-Resolution Image Synthesis},
  author={Podell, Dustin and English, Zion and Lacey, Kyle and Blattmann, Andreas and Dockhorn, Tim and M{\"u}ller, Jonas and Penna, Joe and Rombach, Robin},
  booktitle={ICLR},
year={2024}
}

@inproceedings{yi2024towards,
  title={Towards Understanding the Working Mechanism of Text-to-Image Diffusion Model},
  author={Yi, Mingyang and Li, Aoxue and Xin, Yi and Li, Zhenguo},
  booktitle={NeurIPS},
  year={2024}
}

@inproceedings{somepalli2023diffusion,
  title={Diffusion art or digital forgery? investigating data replication in diffusion models},
  author={Somepalli, Gowthami and Singla, Vasu and Goldblum, Micah and Geiping, Jonas and Goldstein, Tom},
  booktitle={CVPR},
  year={2023}
}

@inproceedings{chen2024towards,
  title={Towards Memorization-Free Diffusion Models},
  author={Chen, Chen and Liu, Daochang and Xu, Chang},
  booktitle={CVPR},
  year={2024}
}

@inproceedings{ross2024geometric,
  title={A Geometric Framework for Understanding Memorization in Generative Models},
  author={Ross, Brendan Leigh and Kamkari, Hamidreza and Liu, Zhaoyan and Wu, Tongzi and Stein, George and Loaiza-Ganem, Gabriel and Cresswell, Jesse C},
  booktitle={ICML 2024 Workshop on GRaM},
  year={2024}
}

@inproceedings{darasconsistent,
  title={Consistent Diffusion Meets Tweedie: Training Exact Ambient Diffusion Models with Noisy Data},
  author={Daras, Giannis and Dimakis, Alex and Daskalakis, Constantinos Costis},
  booktitle={ICML},
  year={2024}
}

@inproceedings{daras2024ambient,
  title={Ambient diffusion: Learning clean distributions from corrupted data},
  author={Daras, Giannis and Shah, Kulin and Dagan, Yuval and Gollakota, Aravind and Dimakis, Alex and Klivans, Adam},
  booktitle={NeurIPS},
  year={2024}
}

@inproceedings{jiang2023ai,
  title={AI Art and its Impact on Artists},
  author={Jiang, Harry H and Brown, Lauren and Cheng, Jessica and Khan, Mehtab and Gupta, Abhishek and Workman, Deja and Hanna, Alex and Flowers, Johnathan and Gebru, Timnit},
  booktitle={AIES},
  year={2023}
}

@misc{orrick2023andersen,
  author = {Orrick, W. H.},
  title = {Andersen v. Stability AI Ltd.},
  year = {2023},
  note = {\url{https://casetext.com/case/andersen-v-stability-ai-ltd}}
}

@inproceedings{hintersdorf2024finding,
  title={Finding NeMo: Localizing Neurons Responsible For Memorization in Diffusion Models},
  author={Hintersdorf, Dominik and Struppek, Lukas and Kersting, Kristian and Dziedzic, Adam and Boenisch, Franziska},
  booktitle={NeurIPS},
  year={2024}
}

@article{kowalczuk2025finding,
  title={Finding Dori: Memorization in Text-to-Image Diffusion Models Is Not Local},
  author={Kowalczuk, Antoni and Hintersdorf, Dominik and Struppek, Lukas and Kersting, Kristian and Dziedzic, Adam and Boenisch, Franziska},
  journal={arXiv preprint arXiv:2507.16880},
  year={2025}
}

@inproceedings{ho2020denoising,
  title={Denoising diffusion probabilistic models},
  author={Ho, Jonathan and Jain, Ajay and Abbeel, Pieter},
  booktitle={NeurIPS},
  year={2020}
}

@inproceedings{songdenoising,
  title={Denoising Diffusion Implicit Models},
  author={Song, Jiaming and Meng, Chenlin and Ermon, Stefano},
  booktitle={ICLR},
  year={2021}
}

@inproceedings{song2021maximum,
  title={Maximum likelihood training of score-based diffusion models},
  author={Song, Yang and Durkan, Conor and Murray, Iain and Ermon, Stefano},
  booktitle={NeurIPS},
  year={2021}
}

@inproceedings{saharia2022photorealistic,
  title={Photorealistic text-to-image diffusion models with deep language understanding},
  author={Saharia, Chitwan and Chan, William and Saxena, Saurabh and Li, Lala and Whang, Jay and Denton, Emily L and Ghasemipour, Kamyar and Gontijo Lopes, Raphael and Karagol Ayan, Burcu and Salimans, Tim and others},
  booktitle={NeurIPS},
  year={2022}
}

@article{hongmembench,
  title={MemBench: Memorized Image Trigger Prompt Dataset for Diffusion Models},
  author={Hong, Chunsan and Oh, Tae-Hyun and Sung, Minhyuk},
  journal={TMLR},
  year={2025}
}

@inproceedings{lin2014microsoft,
  title={Microsoft coco: Common objects in context},
  author={Lin, Tsung-Yi and Maire, Michael and Belongie, Serge and Hays, James and Perona, Pietro and Ramanan, Deva and Doll{\'a}r, Piotr and Zitnick, C Lawrence},
  booktitle={ECCV},
  year={2014},
}

@misc{santana_gustavostastable-diffusion-prompts_2022,
  title = {Gustavosta/{{Stable-Diffusion-Prompts}} {$\cdot$} {{Datasets}} at {{Hugging Face}}},
  author = {Santana, Gustavo},
  year = {2022},
  month = dec,
  url = {https://huggingface.co/datasets/Gustavosta/Stable-Diffusion-Prompts},
  urldate = {2023-01-26}
}

@inproceedings{jeon2025understanding,
title={Understanding and Mitigating Memorization in Generative Models via Sharpness of Probability Landscapes},
author={Dongjae Jeon and Dueun Kim and Albert No},
booktitle={ICML},
year={2025},
url={https://openreview.net/forum?id=EW2JR5aVLm}
}

@article{gumemorization,
  title={On Memorization in Diffusion Models},
  author={Gu, Xiangming and Du, Chao and Pang, Tianyu and Li, Chongxuan and Lin, Min and Wang, Ye},
  journal={TMLR},
  year = {2025}
}

@inproceedings{jiangimage,
  title={Image-level Memorization Detection via Inversion-based Inference Perturbation},
  author={Jiang, Yue and Lin, Haokun and Bai, Yang and Peng, Bo and Liu, Zhili and Lyu, Yueming and Yang, Yong and Dong, Jing and others},
  booktitle={ICLR},
  year={2025}
}

@inproceedings{esser2024scaling,
  title={Scaling Rectified Flow Transformers for High-Resolution Image Synthesis},
  author={Esser, Patrick and Kulal, Sumith and Blattmann, Andreas and Entezari, Rahim and M{\"u}ller, Jonas and Saini, Harry and Levi, Yam and Lorenz, Dominik and Sauer, Axel and Boesel, Frederic and others},
  booktitle={ICML},
  year={2024},
}

@misc{flux2024,
    title={FLUX},
    author={{Black Forest Labs}},
    year={2024},
    howpublished={\url{https://github.com/black-forest-labs/flux}},
}

@article{kim2025diffusion,
  title={How Diffusion Models Memorize},
  author={Kim, Juyeop and Kim, Songkuk and Lee, Jong-Seok},
  journal={arXiv preprint arXiv:2509.25705},
  year={2025}
}

@inproceedings{zhang2018unreasonable,
  title={The unreasonable effectiveness of deep features as a perceptual metric},
  author={Zhang, Richard and Isola, Phillip and Efros, Alexei A and Shechtman, Eli and Wang, Oliver},
  booktitle={CVPR},
  year={2018}
}
